\documentclass[runningheads]{llncs}

\usepackage{eccv}

\usepackage{eccvabbrv}

\usepackage{graphicx}
\usepackage{tabularx}
\usepackage{caption}
\usepackage{booktabs}

\usepackage[accsupp]{axessibility}  %

\usepackage{hyperref}

\usepackage{orcidlink}

\usepackage{booktabs} %
\usepackage{textgreek}
\usepackage[export]{adjustbox}

\newcommand{\mytilde}{\raise.17ex\hbox{$\scriptstyle\mathtt{\sim}$}}

\newcommand{\suppmat}[1]{{\color{black}#1}}

\def\code#1{\texttt{#1}}

\usepackage[super]{nth}

\newcommand{\methodName}{\textPhi{}eat\xspace}

\newcommand{\myparagraph}[1]{\vspace{4pt}\noindent\textbf{#1}}
\begin{document}

\title{\textbf{\methodName:} Physically Grounded Material Feature Representation}

\titlerunning{\textbf{\methodName:} Physically Grounded Material Feature Representation}

\author{Giuseppe Vecchio\orcidlink{0000-0001-5009-4365} \and
Adrien Kaiser\orcidlink{0000-0002-5998-3932} \and
Claudia Cuttano\orcidlink{0009-0004-9672-507X}\and
Romain Rouffet\orcidlink{0009-0001-4412-7017}\and
Rosalie Martin\orcidlink{0009-0002-6812-2664}\and
Elena Garces\orcidlink{0000-0003-3509-8485}\and
Tamy Boubekeur\orcidlink{0000-0001-5985-0921}}

\authorrunning{G.~Vecchio et al.}

\institute{Adobe Research, France\\
\email{\{gvecchio, akaiser, ccuttano, rouffet, rmartin, elenag, boubek\}@adobe.com}}

\maketitle

\begin{figure}[ht]
    \centering
    \begin{minipage}{0.83\textwidth}
        \includegraphics[width=\linewidth]{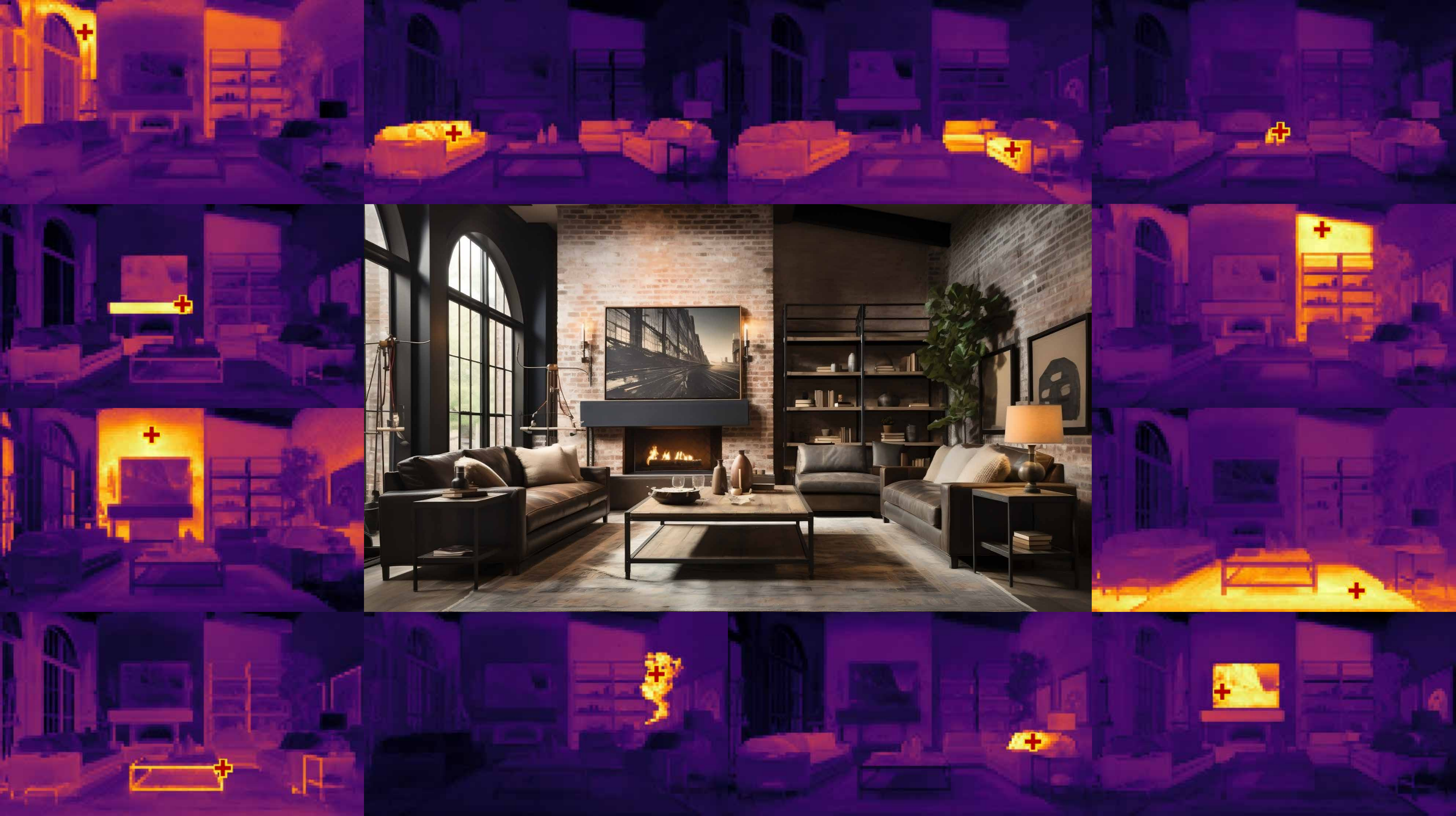}
        \caption{We present \methodName, a novel physically grounded foundation model sensitive to the physical material properties that govern real-world appearance. We visualize the cosine similarity of \methodName\ output features between a query patch, marked with a red cross, and all other patches.}
        \label{fig:teaser}
  \end{minipage}
\end{figure}

\begin{abstract}
While foundation models have emerged as general-purpose visual backbones, their representations are primarily optimized for semantics and lack explicit modeling of physical factors, such as reflectance, hindering their efficacy in tasks requiring explicit material reasoning. 
We introduce \textbf{\methodName}, a novel material-grounded visual backbone that encourages a representation sensitive to material identity, including reflectance and mesostructure. 
Instead of relying on generic data augmentations, we pretrain our model by contrasting observations of the same material under controlled variations in lighting and geometry. This encourages invariance to extrinsic factors while preserving sensitivity to intrinsic material properties.
We show that the resulting representation provides strong priors for material-centric tasks, including feature-based material selection and classification.
Our results demonstrate that physically inspired weak supervision is an effective strategy for learning representations tailored to material perception.
\end{abstract}

\section{Introduction}
\label{sec:introduction}

Modern computer vision has seen a paradigm shift driven by foundation models, such as the DINO family~\cite{caron2021emerging,oquab2023dinov2,simeoni2025dinov3}. These self-supervised models excel at learning representations that are invariant to a wide range of transformations, effectively mapping pixels to a \textit{semantic manifold} where object identity remains stable across different viewpoints and styles. However, while these models capture robust geometric and categorical cues, they often struggle to decouple the physical factors that constitute a scene's appearance, like material properties.

For a representation to truly capture the physical properties that determine the appearance of the world, it must distinguish between \textbf{intrinsic factors}, \ie, the physical material properties an object is made of, and \textbf{extrinsic factors}, \ie, the context in which that material exists, such as the macro-geometry of the object, the global illumination, and the local orientation. In applications like intrinsic decomposition~\cite{garces2022survey}, material capture~\cite{deschaintre2018single}, and robotics, correctly estimating the semantic label (\eg, "chair") is often less critical than the material identity (\eg, "polished oak"). Standard backbones tend to group surfaces by semantic context, whereas material-aware representations should remain stable across changes in shape and lighting for the same intrinsic properties.

The challenge lies in the fact that material identity is not a simple color or texture, but is defined by specific properties, such as reflectance (BRDF), mesostructure, etc.
Standard self-supervised objectives, which rely on photometric augmentations like color jittering or solarization, often inadvertently destroy these subtle physical cues or encourage the model to ignore them in favor of object-level consistency. As a result, specialized tasks requiring deep material reasoning still depend on heavy supervision or narrow, domain-specific datasets.

In this paper, we introduce \textbf{\methodName}, a visual backbone designed to bridge this gap by specializing a pretrained DINOv3 ViT \cite{simeoni2025dinov3} for \textbf{material-aware perception}. 
Our key insight is that instead of generic data augmentations, we can leverage a physically based supervision. By rendering the same material across a diverse set of geometries and lighting environments, we create "physical triplets" that provide a rigorous training signal for intrinsic invariance. We force the model to associate spatial crops of the same material even when their pixel-level appearance changes dramatically due to extrinsic factors. 
To do so, we complement the DINO self-supervised training strategy with a cross-material contrastive loss, which further grounds invariance in real physical properties variation rather than semantic similarity. \methodName shifts the focus of the training from object semantics to intrinsic appearance. This produces features that capture underlying material properties while remaining robust to extrinsic perturbations such as local geometry and complex light interactions.%

The resulting representation is a material-grounded feature extractor that captures intrinsic reflectance and surface structure while remaining robust to the physical context. \methodName provides a powerful prior for downstream tasks, outperforming general-purpose backbones in material-centric applications without requiring dense per-pixel labels during pretraining.

\vspace{15pt}
\noindent In a nutshell, our main contributions are:
\begin{itemize}
    \item \textbf{\methodName}, a weakly supervised visual backbone fine-tuned to encode physically grounded material features such as reflectance and geometric mesostructure;
    \item a material-aware pretraining strategy that leverages synthetic renderings under varied lighting and geometry to encourage invariance to extrinsic appearance factors;
    \item a large-scale, semantically coherent data generation pipeline, leveraging artist-designed mesh templates to render a vast collection of materials.
\end{itemize}

\section{Related Work}
\label{sec:related}

\myparagraph{Intrinsic Scene Understanding.} The problem of recovering and understanding intrinsic scene characteristics, such as surface reflectance, shape, and illumination, from a single image was formalized as early as the 1970s~\cite{land1971lightness,barrow1978recovering}. This is closely related to material perception, where appearance depends on local optical and physical properties rather than object category alone~\cite{adelson2001seeing}. Early work on local visual material attributes further showed that material properties can provide discriminative mid-level cues for recognition beyond object and scene context~\cite{schwartz2015automatically}. Despite decades of progress~\cite{garces2022survey}, intrinsic scene understanding remains only partially solved. Early progress in intrinsic decomposition was driven by non-learning-based approaches, which produced promising results but struggled to generalize. Two main strategies emerged. The first attempted to classify image gradients as either reflectance or shading, relying on heuristics such as entropy minimization~\cite{finlayson2004intrinsic} and Retinex-based assumptions~\cite{land1971lightness,horn1974determining,bi20151}, both of which proved brittle in practice. A second line of work extended this idea by clustering regions of similar reflectance~\cite{garces2012intrinsic,bell2014intrinsic}, typically assuming smooth or continuous lighting within each cluster.
Large-scale material recognition was advanced by the Materials in Context database (MINC)~\cite{bell2015material}, which enabled learning-based material classification and segmentation in real-world images.
Modern learning-based approaches have achieved remarkable progress in intrinsic decomposition~\cite{kocsis2024intrinsic,zeng2024rgbx} and material capture~\cite{deschaintre2018single,martin2022materia}. However, these methods are typically framed as narrow regression tasks. They rely on large-scale, densely labeled datasets and are often restricted to controlled acquisition setups that reduce ambiguity in the estimation, such as flash illumination~\cite{deschaintre2018single,deschaintre2019flexible,gao2019deep,li2018materials} or data captured with dedicated scanning devices~\cite{rodriguez2024textile,garces2023towards}. Some recent approaches attempt to operate under unknown lighting conditions~\cite{vecchio2021surfacenet,martin2022materia,zhou2021adversarial,vecchio2024controlmat}, but they still rely on task-specific supervision.
More recently, material segmentation methods~\cite{sharma2023materialistic,guerrero2025fine} have attempted to leverage the representational power of foundation models such as DINO. SAMa~\cite{fischer2026sama} similarly targets material-aware selection and segmentation, but in 3D assets, using cross-view consistency to propagate material selections across arbitrary 3D representations. These approaches are highly effective for their target tasks, but typically rely on frozen semantic backbones, additional supervised layers, or task-specific selection pipelines to compensate for the lack of physical awareness in the underlying features. This reveals a fundamental limitation: existing representations are primarily organized around semantic similarity, providing only a weak foundation for reasoning about materials. In contrast, \methodName learns generic, material-aware features through physically grounded weak supervision. By encoding intrinsic appearance attributes directly into the backbone, our approach enables robust material reasoning that is not tied to rigid physical priors or frozen semantic representations.

\myparagraph{Foundation Visual Encoders.}
Recent advances in visual representation learning are largely driven by large-scale encoders trained either from image--text pairs or from self-supervision. \emph{Vision--language pretraining} learns transferable representations from large collections of image--text pairs. CLIP~\cite{radford2021clip} aligns image and text embeddings through a contrastive objective. SigLIP replaces the softmax formulation with a pairwise sigmoid loss that improves scalability~\cite{zhai2023siglip}, while SigLIP~2 extends this framework with additional training objectives~\cite{tschannen2025siglip2}. The Perception Encoder (PE)~\cite{bolya2025perception} further shows that contrastive vision--language training yields strong visual representations and exposes different embeddings through alignment strategies: \emph{PE-core} for general-purpose features and \emph{PE Spatial}, which enforces feature consistency within SAM~2~\cite{ravi2024sam} segmentation masks to improve dense prediction. \emph{Self-supervised learning (SSL)} learns representations by enforcing invariance across views of the same image. 
Methods evolved from pretext tasks~\cite{doersch2015unsupervised,noroozi2016unsupervised,pathak2016context,zhang2016colorful} to contrastive learning~\cite{chen2020simple,he2020momentum} and negative-free variants such as SwAV, BYOL, and SimSiam~\cite{caron2020unsupervised,grill2020bootstrap,chen2021exploring}. More recently, RADIOv2.5~\cite{heinrich2025radiov2} proposed a multi-teacher distillation approach to consolidate these disparate strengths into a single "universal" visual backbone. Building on these ideas, DINO~\cite{caron2021emerging} introduced a teacher--student self-distillation framework that learns invariant representations across multiple views of the same image. DINOv2~\cite{oquab2023dinov2} extends this paradigm by combining global self-distillation with masked patch prediction, producing representations that capture both global semantics and dense spatial structure. DINOv3~\cite{simeoni2025dinov3} further scales this approach and introduces stabilization mechanisms such as Gram anchoring to preserve dense feature structure during large-scale training.

\textbf{\methodName} \textit{physically grounds} these representations by leveraging observations of the same material under varying lighting and geometry. This weak supervision redefines the notion of \textit{similarity} during pretraining, aligning representations with meaningful physical appearance cues.

\section{Data Collection and Curation}
\label{sec:data_curation}

\definecolor{TemplateBorder}{HTML}{2F80ED} %
\begin{figure*}[t]
    \centering
    \setlength{\tabcolsep}{0.5pt}

    \setlength{\fboxsep}{0pt}  
    \setlength{\fboxrule}{0.9pt}   

    \begin{tabular}{cccc@{\hspace{6pt}}cccc}
        
        \fcolorbox{TemplateBorder}{white}{\includegraphics[width=0.12\linewidth]{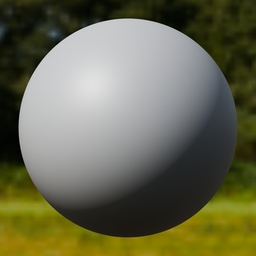}} &
        \includegraphics[width=0.12\linewidth]{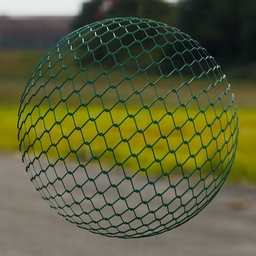} &
        \includegraphics[width=0.12\linewidth]{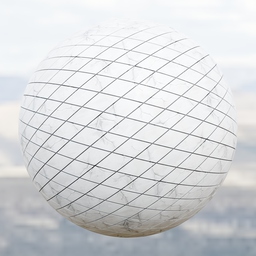} &
        \includegraphics[width=0.12\linewidth]{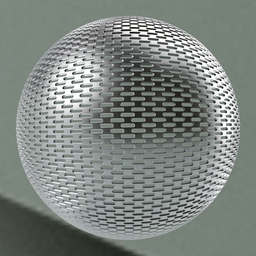} &
        \fcolorbox{TemplateBorder}{white}{\includegraphics[width=0.12\linewidth]{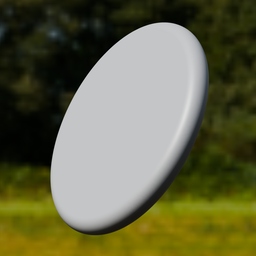}} &
        \includegraphics[width=0.12\linewidth]{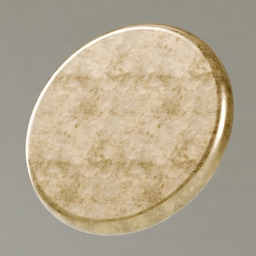} &
        \includegraphics[width=0.12\linewidth]{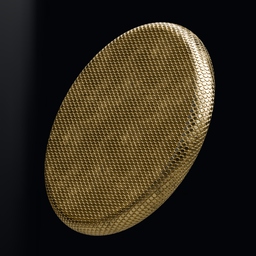} &
        \includegraphics[width=0.12\linewidth]{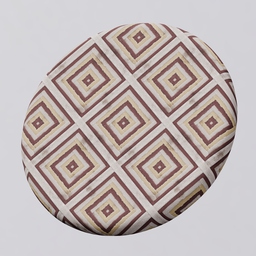} \\

        \fcolorbox{TemplateBorder}{white}{\includegraphics[width=0.12\linewidth]{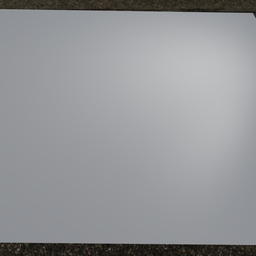}} &
        \includegraphics[width=0.12\linewidth]{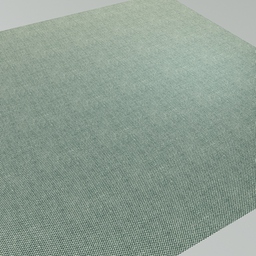} &
        \includegraphics[width=0.12\linewidth]{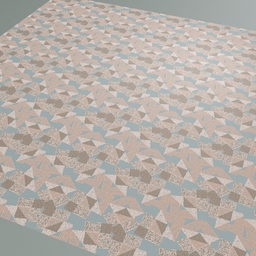} &
        \includegraphics[width=0.12\linewidth]{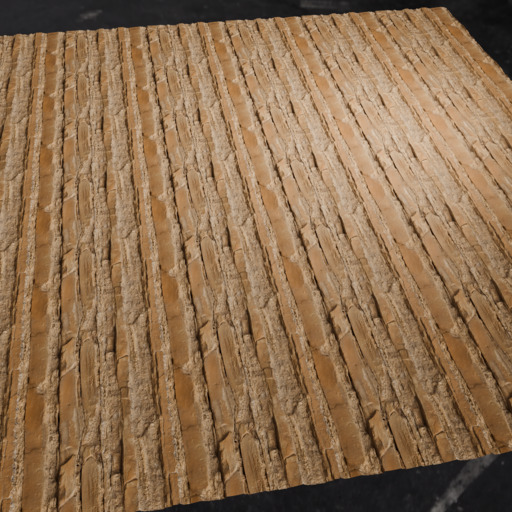} &
        \fcolorbox{TemplateBorder}{white}{\includegraphics[width=0.12\linewidth]{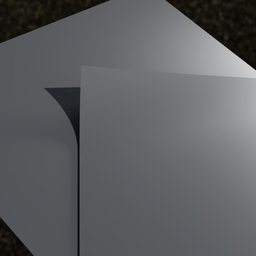}} &
        \includegraphics[width=0.12\linewidth]{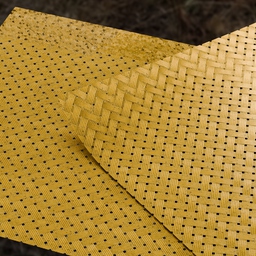} &
        \includegraphics[width=0.12\linewidth]{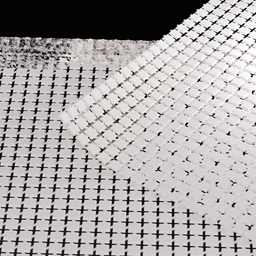} &
        \includegraphics[width=0.12\linewidth]{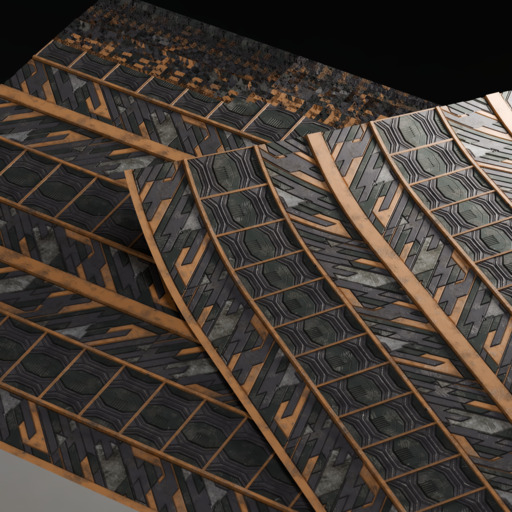} \\

        \fcolorbox{TemplateBorder}{white}{\includegraphics[width=0.12\linewidth]{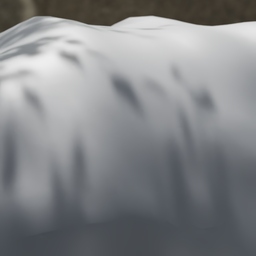}} &
        \includegraphics[width=0.12\linewidth]{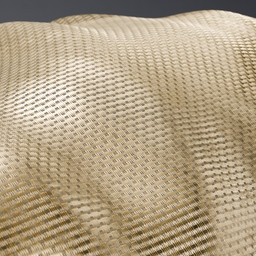} &
        \includegraphics[width=0.12\linewidth]{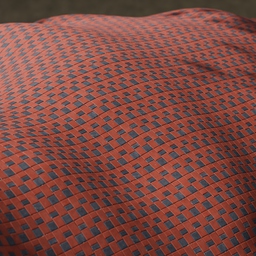} &
        \includegraphics[width=0.12\linewidth]{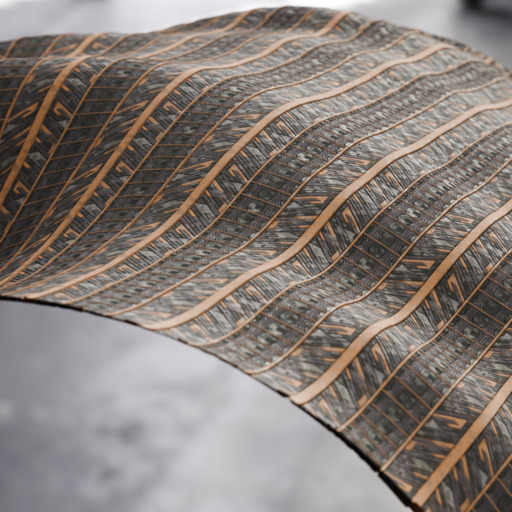} &
        \fcolorbox{TemplateBorder}{white}{\includegraphics[width=0.12\linewidth]{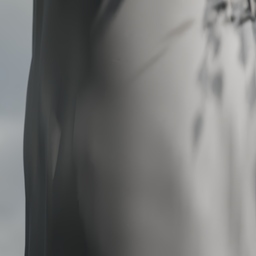}} &
        \includegraphics[width=0.12\linewidth]{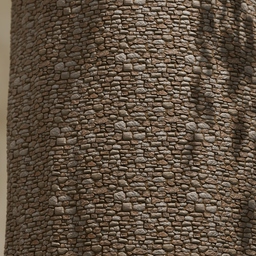} &
        \includegraphics[width=0.12\linewidth]{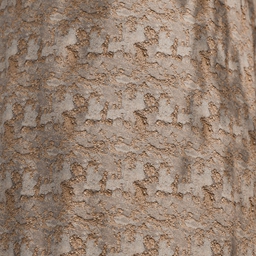} &
        \includegraphics[width=0.12\linewidth]{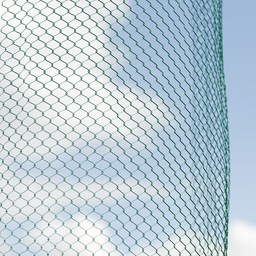} \\
        
    \end{tabular}

    \caption{\textbf{Examples from our curated synthetic material dataset.} Each row shows two geometric templates (highlighted with a blue border) rendered with three different materials each. Templates represent the base geometry onto which materials are applied, while the following images illustrate example renders for different materials on that shape. Templates are semantically aligned with plausible material categories to maintain realistic contexts. The full dataset contains approximately one million high-quality renders generated under diverse lighting, viewpoint, and procedural material variations.}
    \label{fig:dataset}
\end{figure*}

\begin{figure}[ht]
    \centering
    \includegraphics[width=\textwidth]{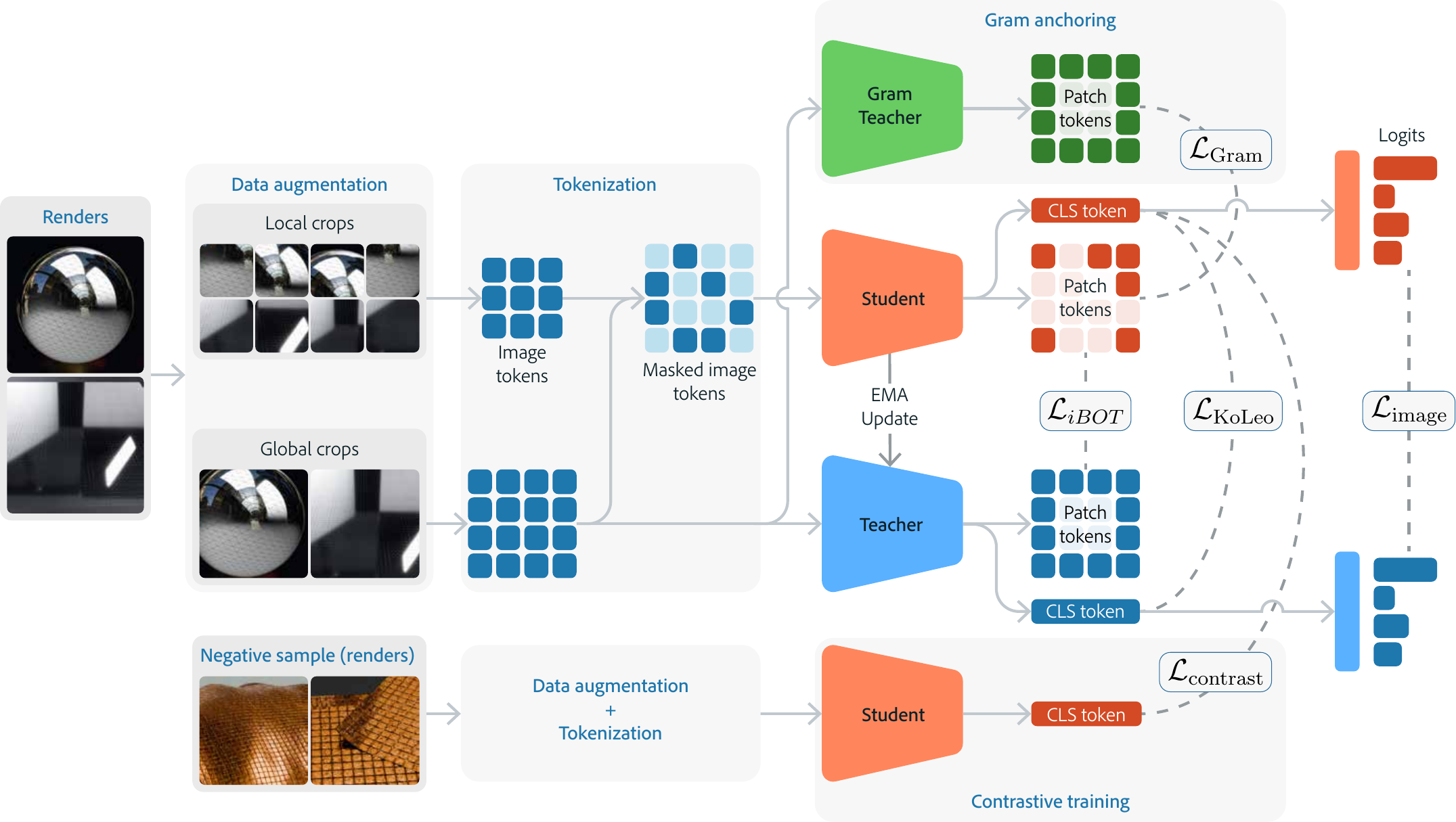}
    \caption{\textbf{\methodName training pipeline.} Two renderings of the same material are sampled and augmented with a multi-crop strategy that yields global and local views. The \textit{student} processes all crops (globals and locals), with random masking on patch tokens for latent reconstruction; the \textit{teacher} and the \textit{Gram teacher} process \emph{global} crops only. Both networks output class- and patch-level embeddings. At the image level, Sinkhorn-balanced teacher assignments supervise student prototype predictions for \emph{all} student views of the same material, defining $\mathcal{L}_{\text{image}}$. At the patch level, masked student tokens are regressed to the teacher tokens at matching spatial indices, giving $\mathcal{L}_{\text{iBOT}}$. On the student global feature \emph{before} the prototype head, KoLeo encourages dispersion ($\mathcal{L}_{\text{KoLeo}}$) and an in-batch InfoNCE pulls together the two views of the same material while pushing away other materials ($\mathcal{L}_{\text{contrast}}$). Gram anchoring aligns second-order structure on global crops. The teacher is an EMA of the student, and the Gram teacher is a frozen snapshot used only for Gram anchoring.}
    \label{fig:architecture}
\end{figure}

The effectiveness of %
foundation model pretraining depends strongly on how the input variability reflects the desired invariances. %
In \methodName, the goal is to disentangle intrinsic physical properties from extrinsic factors such as geometry and illumination. We design a large synthetic dataset which focuses on variability in lighting and shape while keeping the underlying material identity constant. This controlled variability is what enables the model to learn physically grounded features rather than semantic or contextual similarity, albeit without the need to anchor it in a specific physical model \textit{e.g.}, reflectance distribution function.

Similar datasets have been used for inverse rendering~\cite{birsak2025matclip,martin2022materia,vecchio2023matsynth} or material similarity tasks by randomly pairing objects, materials, and lighting. This arbitrary pairing introduces two issues: first, it might bias the training toward unrealistic combinations, affecting material appearance~\cite{Serrano2021_Materials}; second, it makes large-scale generation impractical. %
To address these limitations, we carefully build a set of base geometric templates which we pair with semantically meaningful material categories. For example, a cork material is rendered on a rigid, low-curvature  surface rather than on a highly wrinkled cloth. This semantic alignment between material and macrogeometry allows the network to focus on contexts where those materials are likely to appear in the real world. Fig.~\ref{fig:dataset} shows different pairing examples of templates and materials.

We rely on the Adobe Substance 3D Assets library~\cite{substanceassets2025}, which provides over 9,500 procedural materials covering a wide range of 21 appearance classes, including fabric, metal, wood, stone, marble, and plastic. We vary the procedural parameters of these materials following artist-designed presets, to synthesize approximately 36,000 unique sets of PBR (Physically Based Rendering) maps~\cite{Burley2015ExtendingDisneyBRDF}.
Each material instance is rendered on a collection of semantically aligned geometries, and illuminated under four high dynamic range environment maps randomly selected from a list of 20, providing diverse lighting conditions. \suppmat{The supplementary materials shows additional materials and the environment maps.}

For each render, we randomly vary both the object and environment rotations to ensure a rich sampling of view and light directions. Rendering is performed with a Monte Carlo path tracer using physically accurate light transport and microfacet materials~\cite{Trowbridge1975,walter2007microfacet}, ensuring high realism and consistent energy conservation across all scenes.
We employ GPU-native path tracing with 128 samples per pixel and apply denoising to all renders. We tessellate each object using the material’s displacement map to synthesize realistic self-shadows and inter-reflections. 
We render the environment map behind transparent regions (holes) to increase variability, helping the model separate thin structures from their surroundings.
This process yields roughly one million high-quality renders. %

\section{Method}
\label{sec:method}

\methodName aims to learn visual representations grounded in the intrinsic physical properties of materials. Inspired by recent self-supervised models~\cite{oquab2023dinov2,simeoni2025dinov3} that rely on image-space augmentations to achieve semantic invariance, we propose a pretraining strategy centered on \textbf{physical invariance}. We replace generic photometric jitter with structured variations of extrinsic factors—rendering the same material across diverse geometries and lighting conditions (Fig.~\ref{fig:architecture}). By contrasting these physically grounded views while keeping intrinsic properties constant, \methodName learns to ignore environmental context in favor of material identity.

\subsection{Model Architecture}

\methodName builds upon the vision transformer (ViT) architecture used in DINOv3~\cite{simeoni2025dinov3}. The network consists of a stack of transformer encoder blocks operating on non-overlapping image patches. Each input image is divided into fixed-size patches of $16\times16$ pixels, which are flattened and linearly projected into patch tokens. A special global token (often referred to as the \code{[CLS]} token) is prepended for image-level representation. The model incorporates a small set of learnable ``register tokens'' that act as dedicated memory slots for global context and mitigate patch-token artifacts.
Positional information in the token embeddings is handled via the Rotary Positional Embedding (RoPE)~\cite{su2024roformer} mechanism, which supports token-sequence extrapolation and variable input resolutions. 
The tokens (global, register, patch) are fed into the transformer encoder stack, which outputs one vector for each token. From the final layer we extract both the patch-level features (one per patch) and the global representation vector from the \code{[CLS]} token.

\subsection{Training Objectives}

\methodName follows the self-supervised formulation of DINOv3~\cite{simeoni2025dinov3}, extending it with weak supervision on unlabelled physically varying input pairs.
Each batch contains 2 global crops and 8 local crops extracted from $N$ renderings $(x_1, ..., x_N)$ of the \textbf{same material under different lighting and geometry}. For each crop, the network produces both global and patch-level embeddings. 
Following DINOv3~\cite{simeoni2025dinov3}, we use a combination of image-level objective $\mathcal{L}_{image}$ and a patch-level latent reconstruction objective~\cite{zhou2021ibot} $\mathcal{L}_{iBOT}$, as well as a $\mathcal{L}_{KoLeo}$ regularizer and Gram anchoring. We complement these objectives by introducing an \textbf{in-batch contrastive term} which pulls together representations of the same material while pushing away the embeddings corresponding to different materials.
This combination of objectives encourages alignment between physically equivalent renderings while maintaining diversity across unrelated materials.

\myparagraph{Image-level objective.}
The global alignment loss follows the formulation of the DINOv3 objective~\cite{simeoni2025dinov3}, which replaces the centering and sharpening mechanisms of DINOv2 with the Sinkhorn–Knopp normalization from SwAV~\cite{caron2020unsupervised}.
Such normalization of the teacher probabilities enforces balanced assignments over $K$ learnable prototypes, improving stability and preventing feature collapse. Let $f_s(v)$ and $f_t(v)$ be the student and teacher global embeddings for view $v$, and let $W = [w_1,\dots,w_K]^\top \in \mathbb{R}^{K \times D}$ be the learnable prototype matrix. 
Student probabilities are
\[
p_k(v_s) \;=\; \frac{\exp\!\big(\tfrac{1}{\tau_s}\, w_k^\top f_s(v_s)\big)}{\sum_{\ell=1}^{K}\exp\!\big(\tfrac{1}{\tau_s}\, w_\ell^\top f_s(v_s)\big)} ,
\]
while teacher assignments $q_k(v_t)$ are obtained by applying Sinkhorn–Knopp normalization~\cite{caron2020unsupervised} to the teacher logits $\tfrac{1}{\tau_t}W f_t(v_t)$ over the batch, enforcing balanced use of the $K$ prototypes. 
Given a set $V_t$ of teacher views and $V_s$ of student views, the cross-entropy objective is
\begin{equation}
\label{eq:dino_image}
\mathcal{L}_{\text{image}}
\;=\;
-\frac{1}{|V_t|\,|V_s|}
\sum_{v_t \in V_t}\sum_{v_s \in V_s}
\sum_{k=1}^{K} q_k(v_t)\,\log p_k(v_s).
\end{equation}

\myparagraph{Patch-level latent objective.}
Following iBOT~\cite{zhou2021ibot} and DINOv3~\cite{simeoni2025dinov3}, a masked latent reconstruction objective is applied at the patch level. A random subset of the student’s patch tokens is masked out, and the network is trained to predict the corresponding patch embeddings produced by the teacher on the same spatial positions:
\begin{equation}
    \mathcal{L}_{iBOT} = \frac{1}{M} \sum_{m=1}^{M} 
    \left\| h_\theta(p^s_m) - p^t_m \right\|_2^2,
\end{equation}
where $p^s_m$ and $p^t_m$ denote student and teacher patch embeddings, $h_\theta$ is a projection head, and $M$ is the number of masked patches.

\myparagraph{KoLeo loss.}
We use the Kozachenko–Leonenko entropy estimator on the batch of $L_2$-normalized student features to encourage dispersion and avoid collapse. 
Let $\{z_i\}_{i=1}^{B}$ be the normalized global student embeddings in a batch and let $\rho_i = \min_{j\neq i}\|z_i - z_j\|_2$ be the nearest-neighbor distance for sample $i$. 
Ignoring constants independent of the parameters, maximizing entropy corresponds to minimizing
\begin{equation}
\label{eq:koleo}
\mathcal{L}_{\text{KoLeo}}
\;=\;
-\frac{1}{B}\sum_{i=1}^{B} \log\big(\rho_i + \varepsilon\big),
\end{equation}
with a small $\varepsilon>0$ for numerical stability. 
This promotes a more uniform feature distribution on the unit hypersphere.

\myparagraph{Gram anchoring.}
A secondary frozen teacher (Gram teacher) provides a structural target at patch level. 
We denote by $P_s \in \mathbb{R}^{N \times D}$ and $P_G \in \mathbb{R}^{N \times D}$ the student and Gram-teacher patch matrices after mean-centering across patches. 
We align second-order patch relations via
\begin{equation}
\label{eq:gram}
\mathcal{L}_{\text{Gram}}
\;=\;
\frac{1}{N^2}\,\big\|\, P_s P_s^{\top} - P_G P_G^{\top} \big\|_F^2 .
\end{equation}

\myparagraph{Contrastive loss.}
We complement the teacher–student alignment with an in-batch InfoNCE~\cite{oord2018representation} loss on $L_2$-normalized global student embeddings. 
For each anchor $z_i$, positives $P(i)$ are the \textit{other views of the same material}, and all remaining samples in the batch are negatives:
\begin{equation}
\label{eq:contrast}
\mathcal{L}_{\text{contrast}}
=
-\frac{1}{N}
\sum_{i=1}^{N}
\log
\frac{\sum_{j \in P(i)} \exp\!\big(\operatorname{sim}(z_i, z_j)/\tau\big)}
{\sum_{k \neq i} \exp\!\big(\operatorname{sim}(z_i, z_k)/\tau\big)} ,
\end{equation}
where $\operatorname{sim}$ is cosine similarity and $\tau$ is a temperature.
This objective explicitly pulls together representations of the same physical material, while pushing apart embeddings corresponding to different materials. It complements the DINO objective by reinforcing instance-level consistency independently of the teacher distribution, anchoring the model’s global features to material identity.

\myparagraph{Total objective.}
The total loss consists of a weighted sum of these components:
\begin{equation}
\label{eq:total}
    \mathcal{L}_{\text{total}}
    =
    \mathcal{L}_{\text{image}}
    + \lambda_{\text{p}}\,\mathcal{L}_{iBOT}
    + \lambda_{\text{k}}\,\mathcal{L}_{\text{KoLeo}}
    + \lambda_{\text{g}}\,\mathcal{L}_{\text{Gram}}
    + \lambda_{\text{c}}\,\mathcal{L}_{\text{contrast}} ,
\end{equation}
with the coefficients reported in Section~\ref{sec:training}.

The combination of our augmentation strategy and the cross-material contrastive loss shifts the inductive bias of the pretraining from semantic consistency across image crops to physical invariance across varying extrinsic conditions.

\section{Experiments}
\label{sec:results}

We evaluate whether \methodName learns representations that are invariant to extrinsic factors, such as lighting and geometry, while remaining sensitive to intrinsic material properties. We consider three main evaluations: \textbf{material selection}, based on patch-level material similarity; \textbf{feature separability}, measured with a non-parametric $k$-nearest neighbors (k-NN) classifier; and \textbf{robustness}, measured through prediction stability under illumination and geometry changes. Additionally, we assess the \textbf{semantic--material tradeoff} by evaluating how much general-purpose semantic performance is retained after material-centric adaptation. Together, these experiments test whether the learned features capture material identity independently of environmental conditions and quantify the compromise on semantic capabilities.

\myparagraph{Baselines.}
We compare \methodName against strong vision foundation models trained with different supervision paradigms. We consider CLIP~\cite{radford2021clip} and SigLIP~2~\cite{tschannen2025siglip2} as vision–language models trained on large-scale image–text pairs, the Perception Encoder~\cite{bolya2025perception} in both its PE-core and PE Spatial variants, the agglomerative RADIOv3~\cite{heinrich2025radiov2}, and the self-supervised models DINOv2 with registers~\cite{darcet2023vision, oquab2023dinov2} and DINOv3~\cite{simeoni2025dinov3}. These models represent the strongest publicly available visual encoders. All methods use their \textit{ViT-B} backbones, and we adopt input resolutions corresponding to $1024$ patch tokens ($448\times448$ for patch size $14$, $512\times512$ for patch size $16$) for a fair comparison.

\myparagraph{Implementation Details.}
\label{sec:training} We train \methodName for 10,000 iterations using batches of $512$ render pairs, each consisting of two different views of the same material. Each pair is augmented using a multi-crop strategy, producing 2 global and 8 local crops per view. The global and local views are extracted by randomly cropping 40–100\% and 10–40\% of the renders respectively, ensuring a diverse distribution of context and texture. We use a \textit{ViT-B/16} backbone initialized from DINOv3 weights and optimized using AdamW~\cite{loshchilov2018adamw} with a base learning rate of $0.001$ and weight decay of $0.05$. The teacher momentum follows a cosine schedule between $0.996$ and $1.0$, and the student temperature $\tau_s$ is fixed to $0.1$. We train on input resolutions of $224^2$ for global crops and $112^2$ for local crops, following the multi-crop strategy described in Section~\ref{sec:method}. Training is performed on 16 NVIDIA A100 GPUs using mixed-precision distributed data parallelism, for a total approximate time of 100 hours. We set the loss weights to $\lambda_{\text{p}}=1.0$, $\lambda_{\text{k}}=0.1$, $\lambda_{\text{g}}=0.7$, and $\lambda_{\text{c}}=0.25$. For the KoLeo term we use $\varepsilon=10^{-6}$ in Eq.~\eqref{eq:koleo}; for the patch objective we randomly mask between $10\%$ and $50\%$ of the student patches with a probability of $50\%$; the InfoNCE temperature is fixed to $\tau=0.1$. We enable the Gram anchoring for the last 2,000 iteration steps.

\subsection{Quantitative Results}
\label{sec:quant_esults}

\begin{table}[t]
\centering
\setlength{\tabcolsep}{3pt}
\caption{\textbf{Material selection and classification results}.
We compare \methodName{} with state-of-the-art foundation visual encoders trained with different supervision signals.
Pixel-level metrics evaluate material selection performance, while classification metrics assess feature separability using k-NN retrieval.
\methodName{} achieves the strongest overall performance, improving material selection and classification across metrics, indicating that the learned representation is both physically grounded and discriminative.
Best results \textbf{bold}, \nth{2} best \underline{underlined}. Task-specific supervised material selection methods are reported as supervised references.}
\label{tab:selection}

\begin{tabular*}{\linewidth}{@{\extracolsep{\fill}} l ccc | ccc}
\toprule

\textbf{Model}
& \multicolumn{3}{c|}{Material Selection}
& \multicolumn{3}{c}{k-NN Classification} \\
\cmidrule(lr){2-4}\cmidrule(lr){5-7}

& $\ell_1$ $\downarrow$
& mIoU $\uparrow$
& F1 $\uparrow$
& Acc. $\uparrow$
& Prec. $\uparrow$
& F1 $\uparrow$ \\
\midrule

\textit{Supervised (upper bound)} \\
~Materialistic~\cite{sharma2023materialistic}& 5.7 & 85.8 & 90.6 & --- & --- & --- \\
~Guerrero et al. 2025~\cite{guerrero2025fine} & 3.0 & 89.6 & 93.5 & --- & --- & --- \\

\midrule
\textit{Weakly supervised} \\
~CLIP~\cite{radford2021clip} & 55.7 & 21.6 & 31.6 & 39.9 & 36.6 & 31.2  \\
~SigLIP 2~\cite{tschannen2025siglip2} & 53.1 & 24.4 & 35.2 & 33.5 & 30.8 & 25.1  \\
~PE-core~\cite{bolya2025perception} & 37.3 & 40.5 & 54.6 & 44.2 & 37.0 & 33.9  \\

\midrule
\textit{Agglomerative} \\

~PE Spatial~\cite{bolya2025perception} & \textbf{25.2} & {61.4} & 72.3 & 50.3 & 44.0 & 39.0 \\
~RADIOv3~\cite{heinrich2025radiov2} & 25.6 & \underline{63.5} & \underline{74.7} & 35.9 & 32.9 & 28.8 \\

\midrule
\textit{Self-supervised} \\
~DINOv2~\cite{oquab2023dinov2} & 26.7 & 56.5 & 69.3 & 64.3 & 56.9 & 53.3  \\
~DINOv3~\cite{simeoni2025dinov3} & 27.6 & 59.7 & 72.2 & \underline{66.9} & \underline{60.0} & \underline{55.4}  \\

\midrule
~\textbf{\methodName (ours)} & \underline{25.4} & \textbf{72.0} & \textbf{80.8} & \textbf{71.8} & \textbf{75.0} & \textbf{66.6}  \\

\bottomrule
\end{tabular*}
\end{table}

\begin{figure}[t]
    \centering
    \setlength{\tabcolsep}{.5pt}
    \begin{tabular}{ccccccccc}
        && \multicolumn{3}{c}{\textbf{Patch-wise similarity}} & & \multicolumn{3}{c}{\textbf{Unsupervised segmentation}} \\
        Input & \hspace{10pt} & PE Spatial & DINOv3 & \methodName & \hspace{10pt} & PE Spatial & DINOv3 & \methodName \\

        \vspace{-0.8mm}\includegraphics[width=0.131\linewidth]{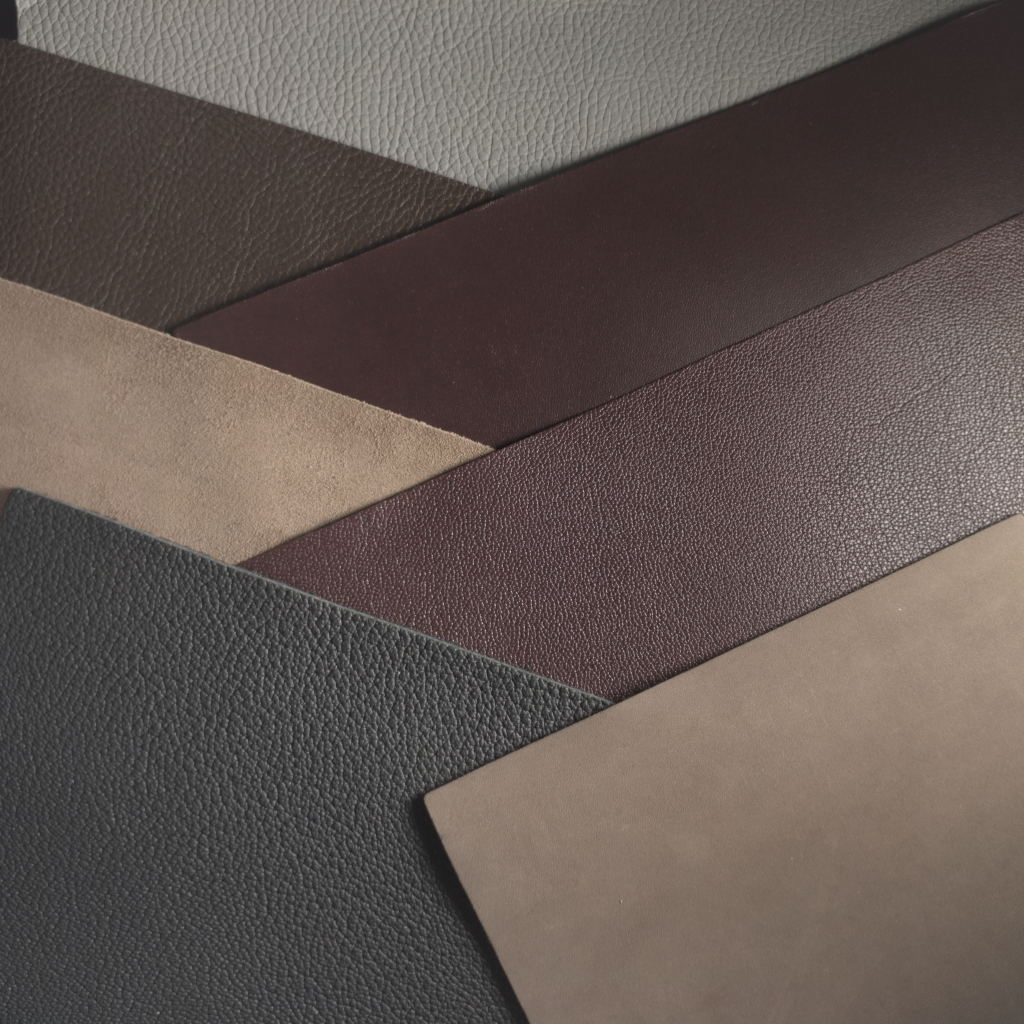} & &
        \includegraphics[width=0.131\linewidth]{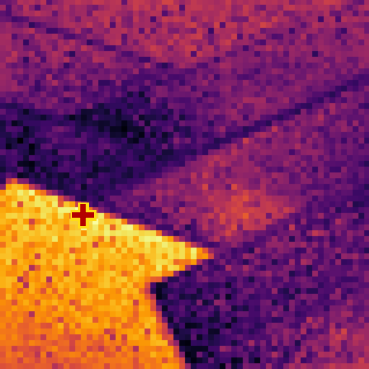} &
        \includegraphics[width=0.131\linewidth]{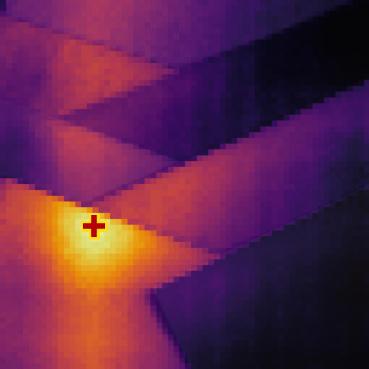} &
        \includegraphics[width=0.131\linewidth]{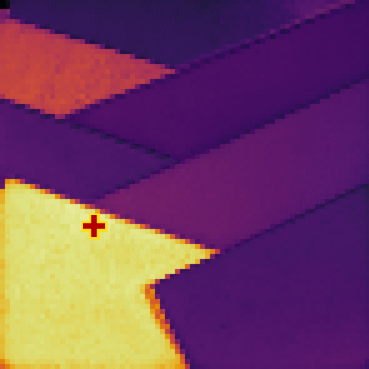} & &
        \includegraphics[width=0.131\linewidth]{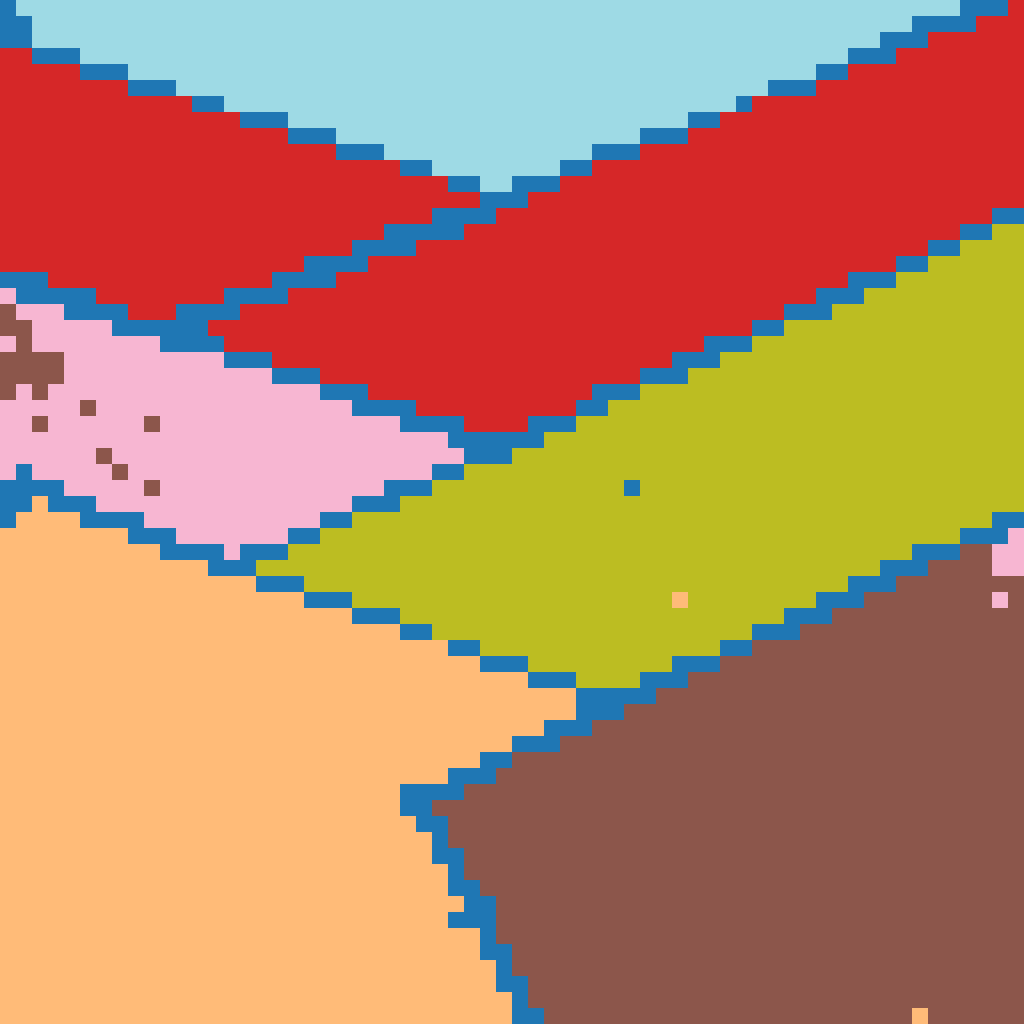} &
        \includegraphics[width=0.131\linewidth]{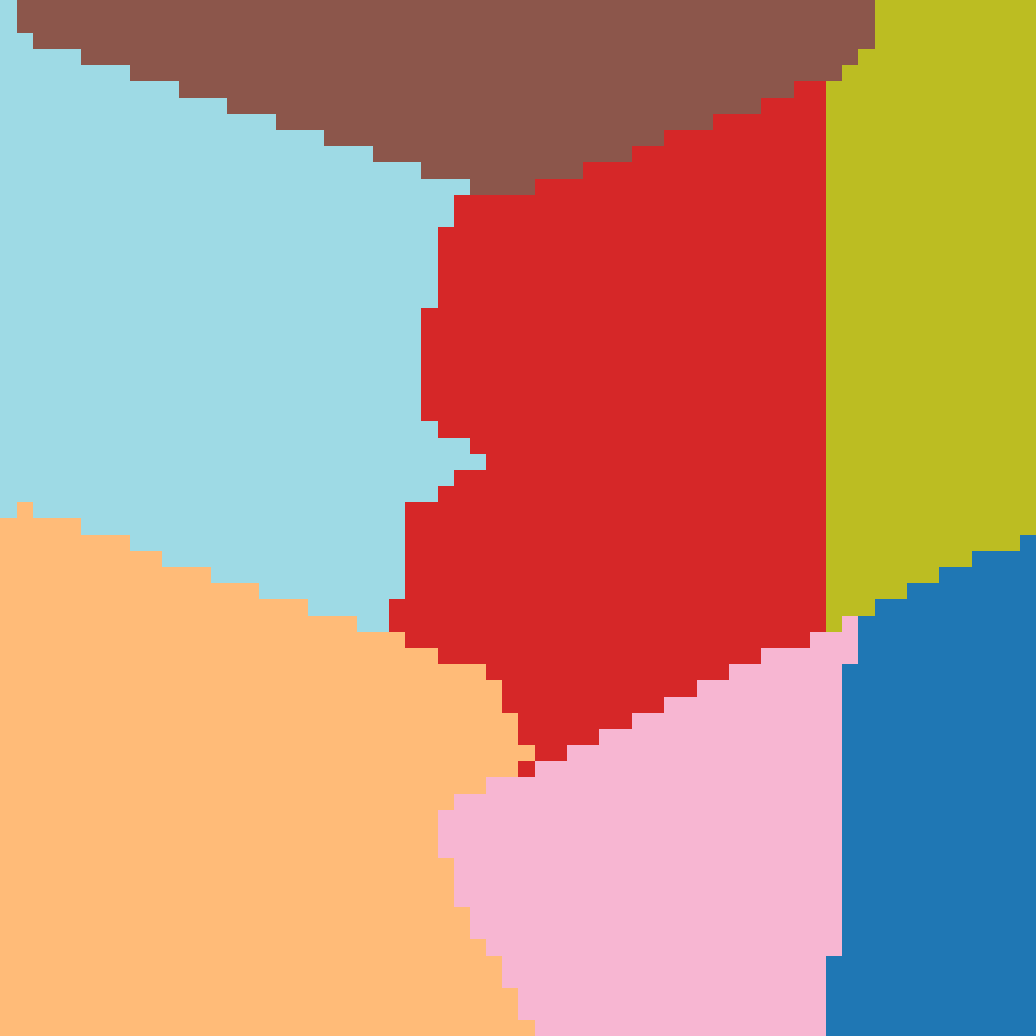} &
        \includegraphics[width=0.131\linewidth]{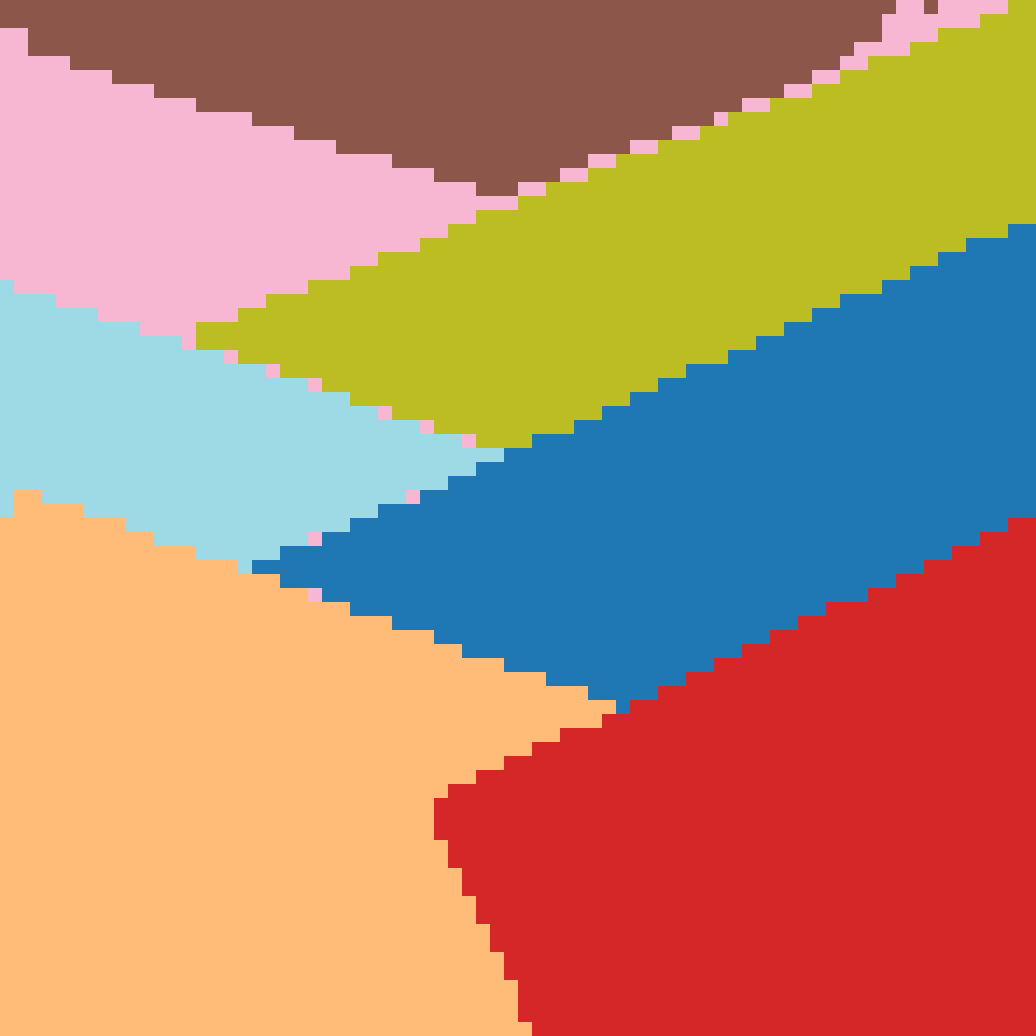} \\
        
        \vspace{-0.8mm}\includegraphics[width=0.131\linewidth]{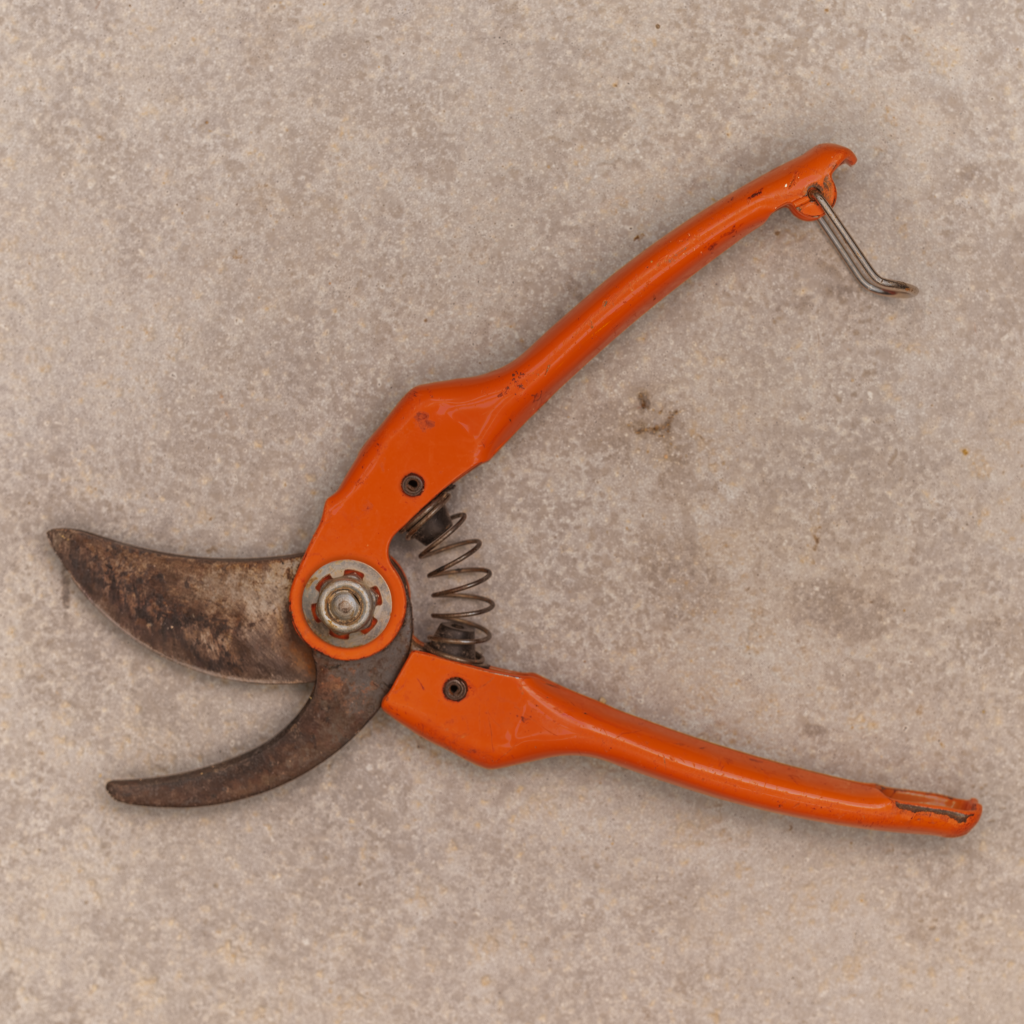} & &
        \includegraphics[width=0.131\linewidth]{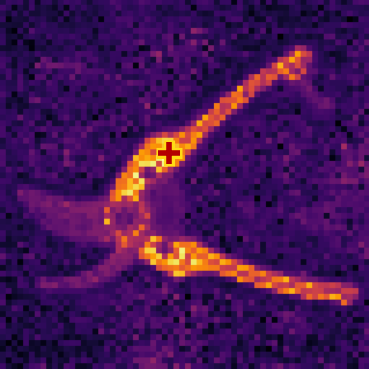} &
        \includegraphics[width=0.131\linewidth]{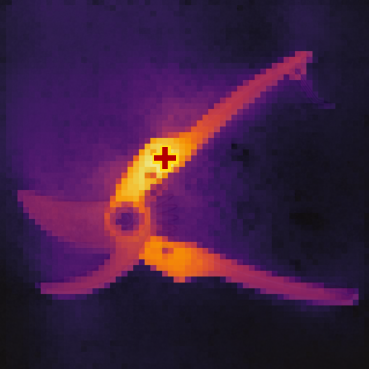} &
        \includegraphics[width=0.131\linewidth]{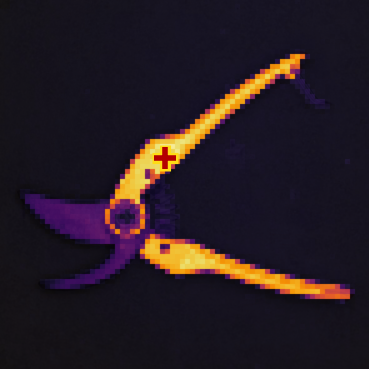} & &
        \includegraphics[width=0.131\linewidth]{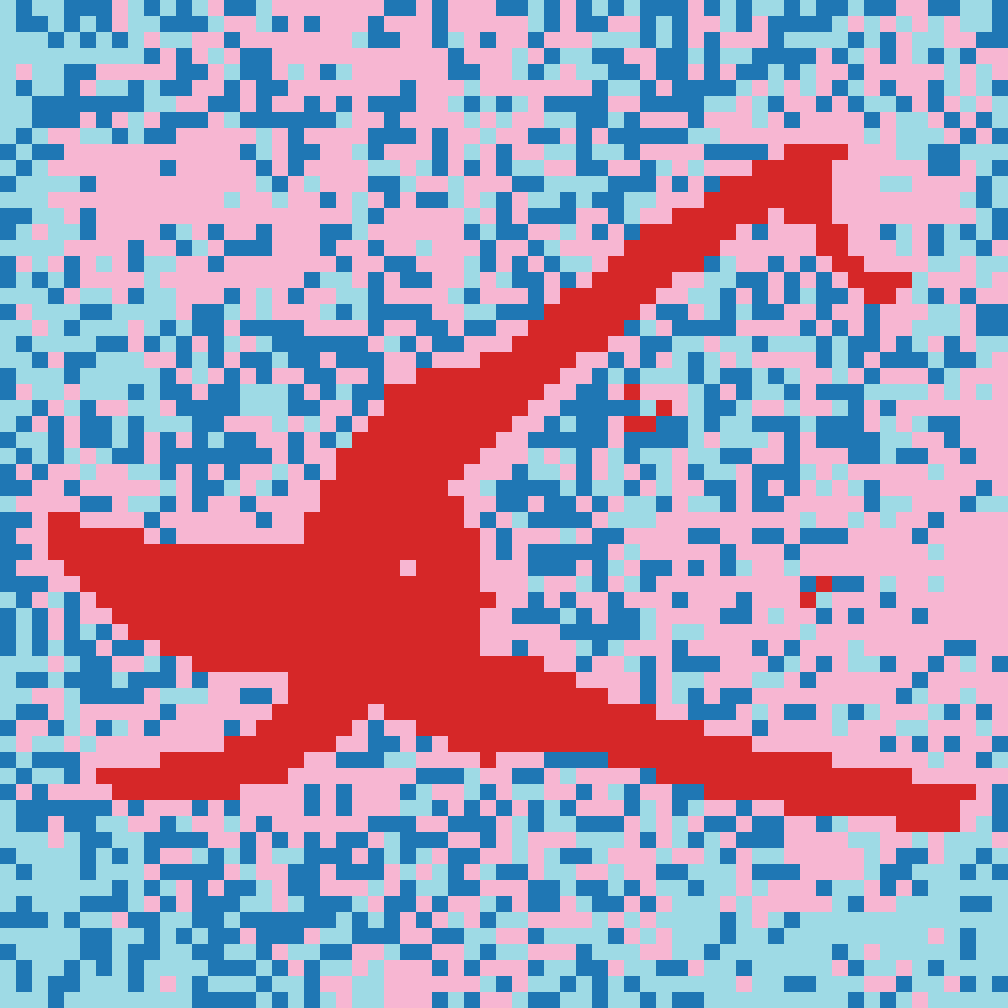} &
        \includegraphics[width=0.131\linewidth]{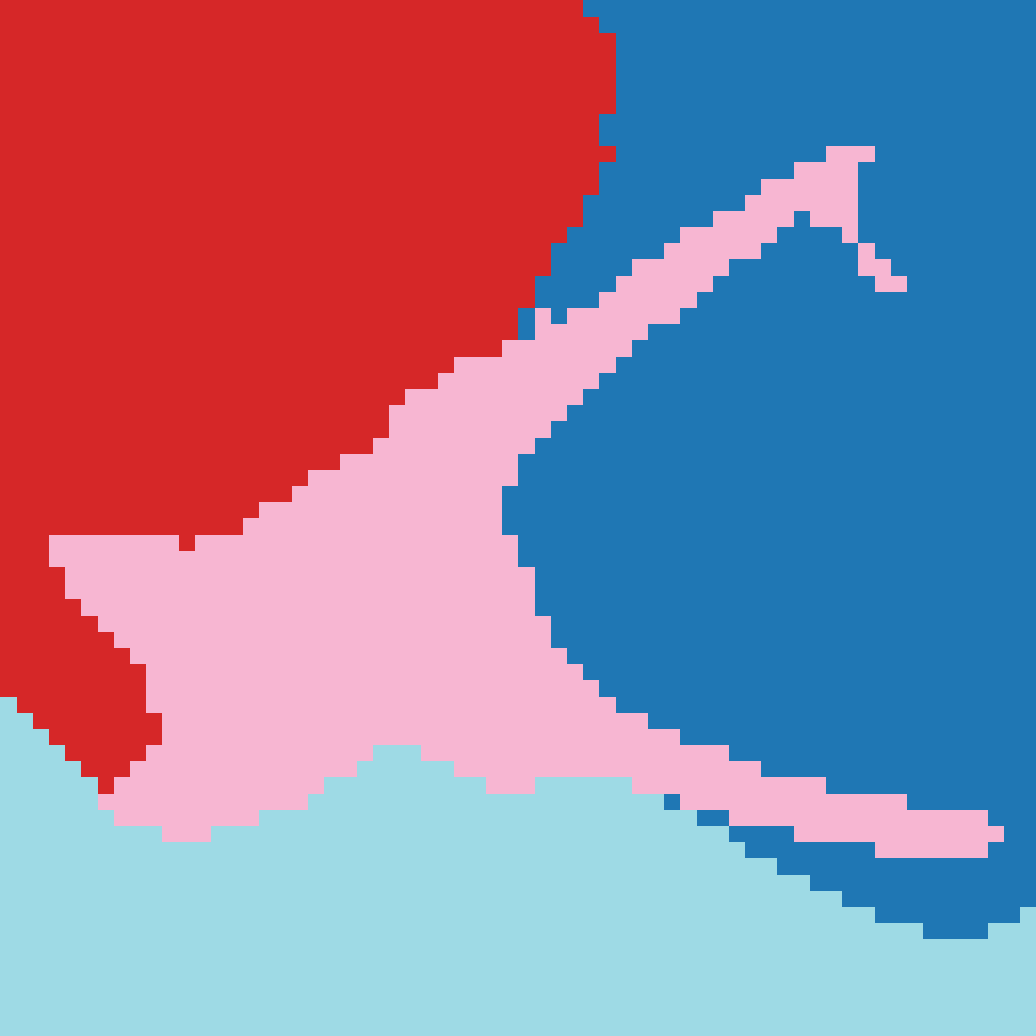} &
        \includegraphics[width=0.131\linewidth]{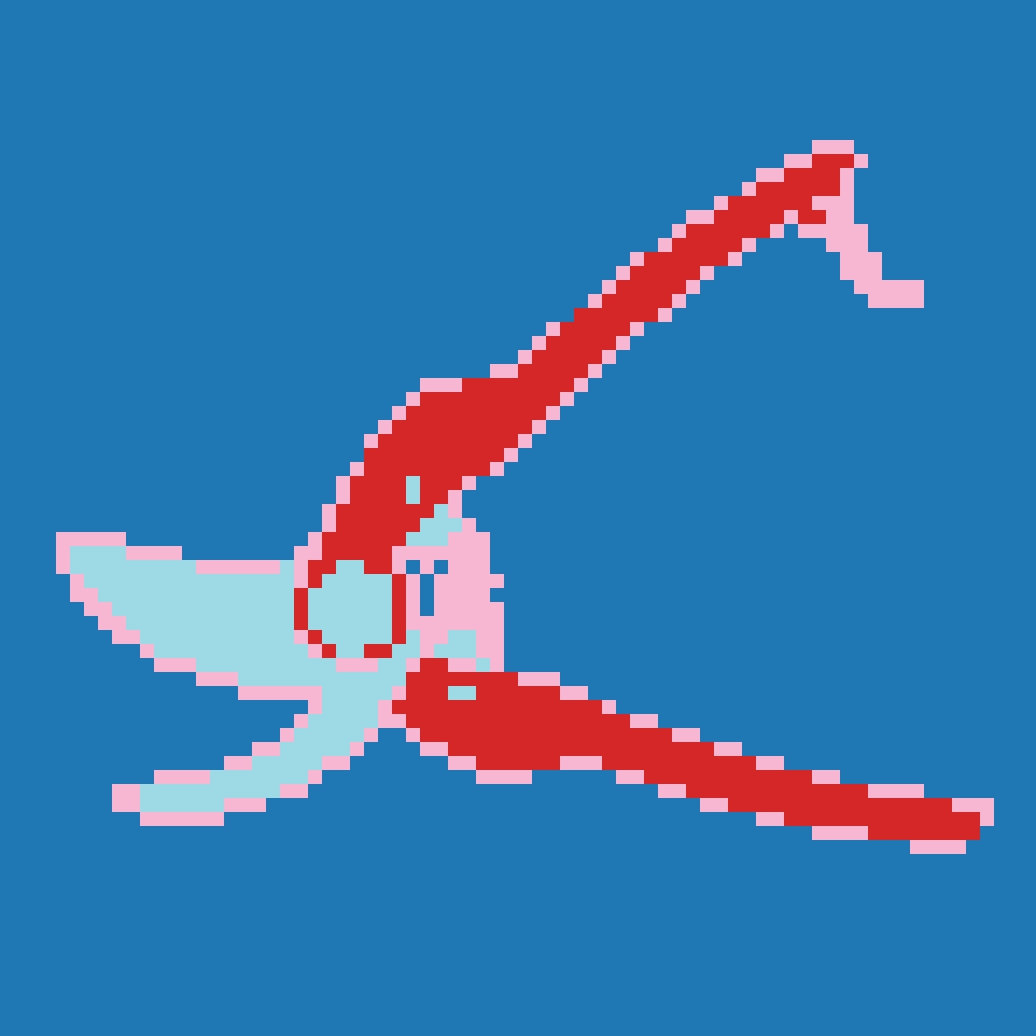} \\

        \vspace{-0.8mm}\includegraphics[width=0.131\linewidth]{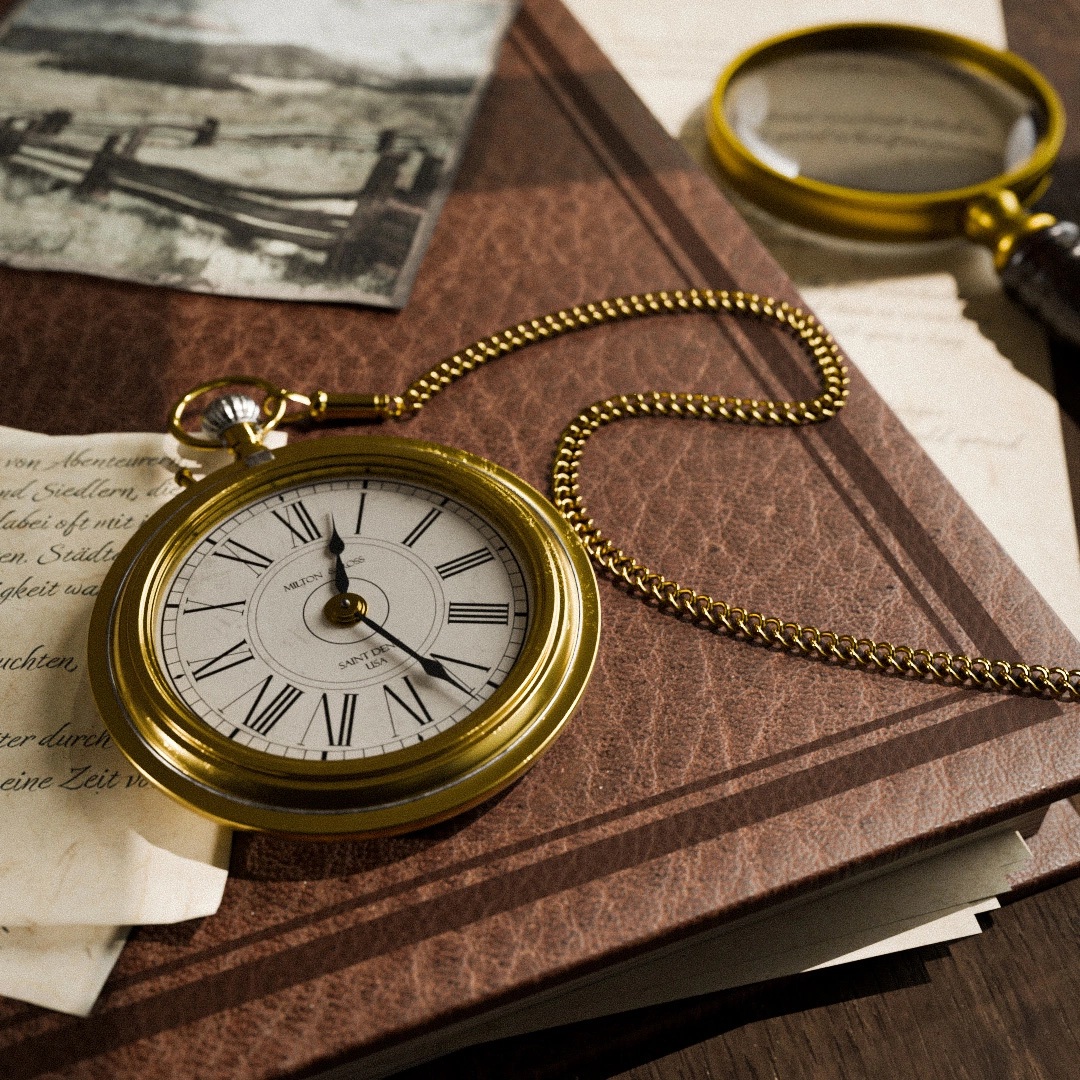} & &
        \includegraphics[width=0.131\linewidth]{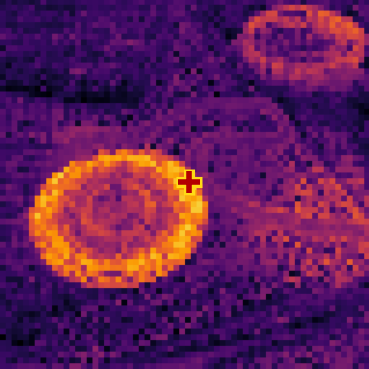} &
        \includegraphics[width=0.131\linewidth]{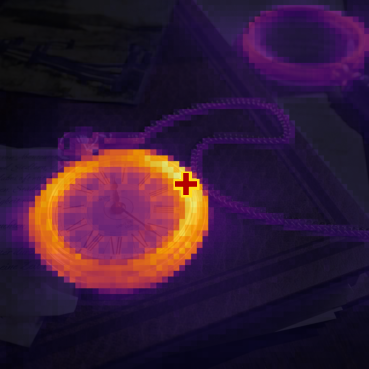} &
        \includegraphics[width=0.131\linewidth]{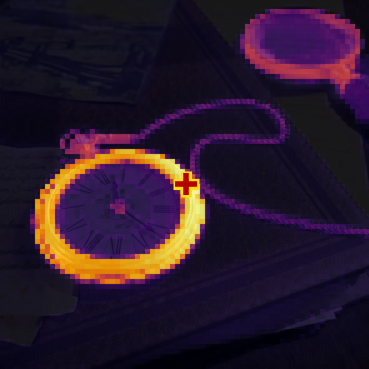} & &
        \includegraphics[width=0.131\linewidth]{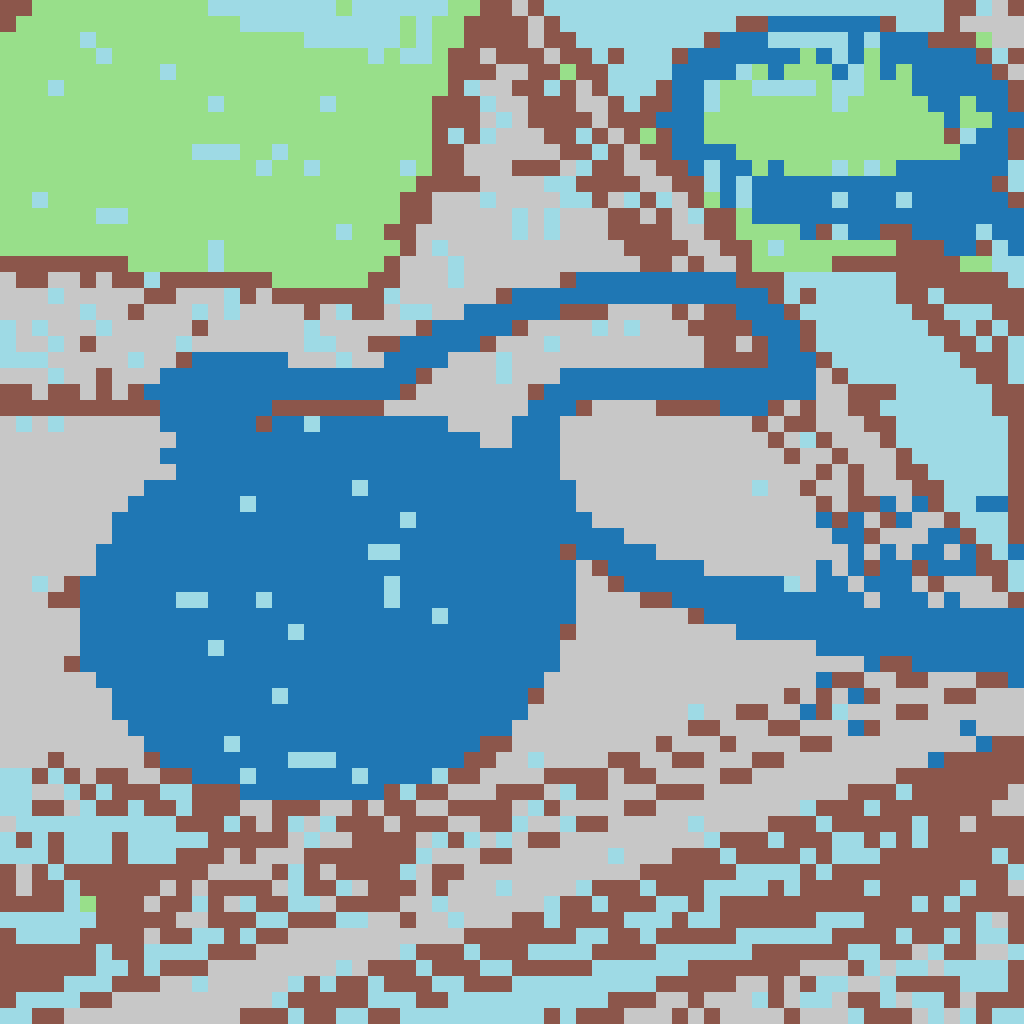} &
        \includegraphics[width=0.131\linewidth]{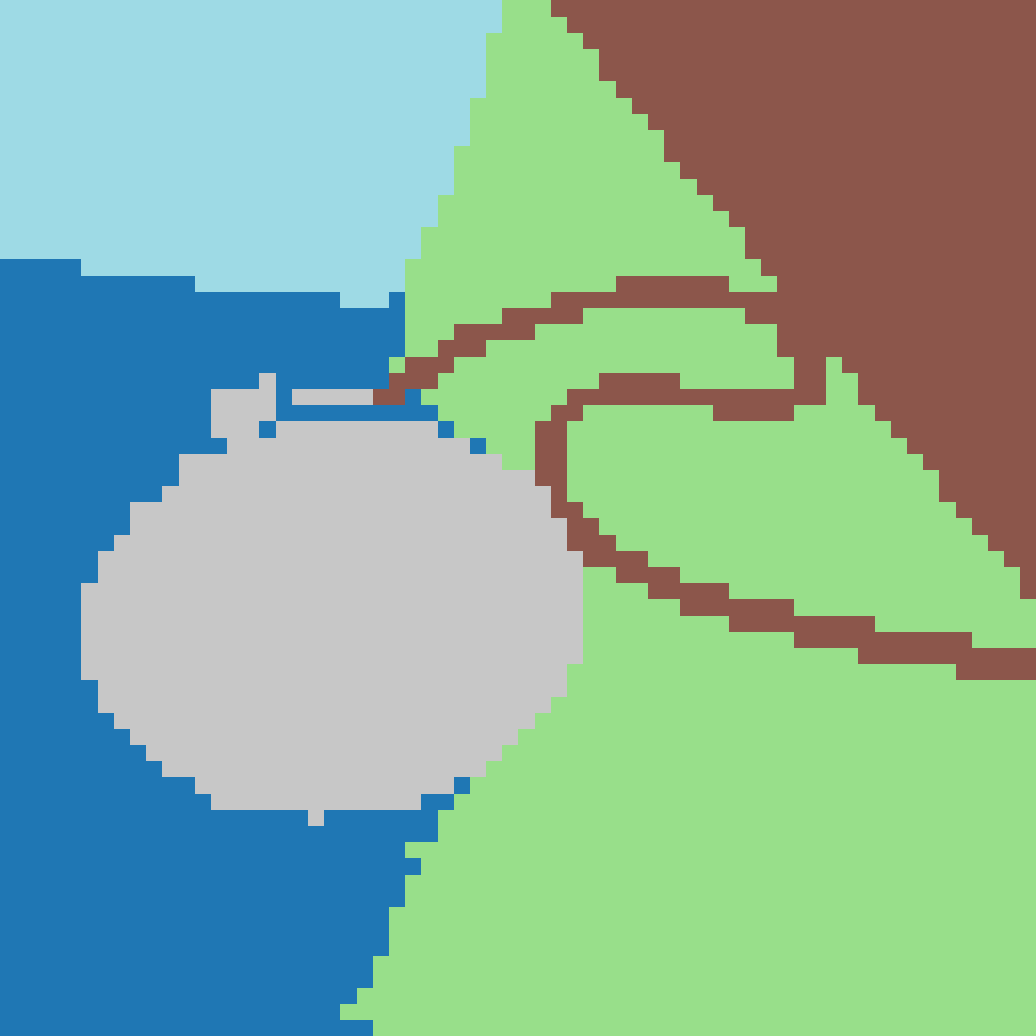} &
        \includegraphics[width=0.131\linewidth]{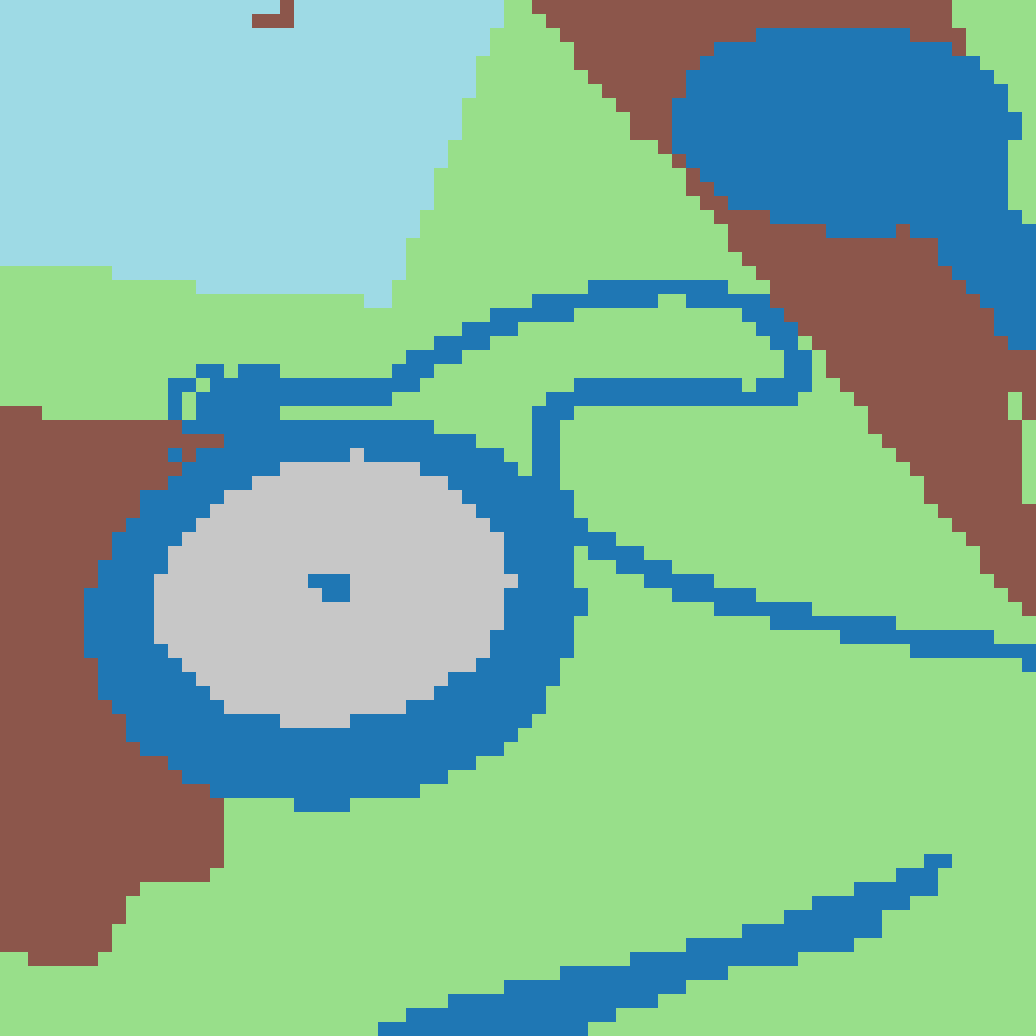} \\

        \vspace{-0.8mm}\includegraphics[width=0.131\linewidth]{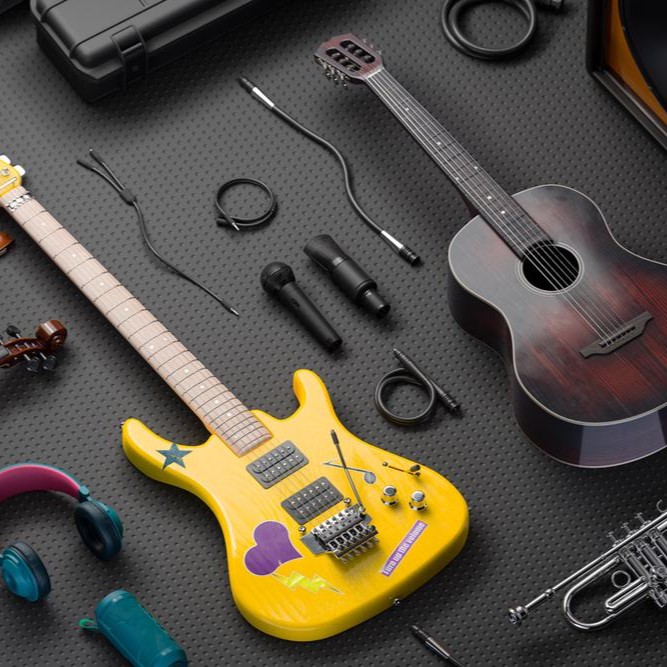} & &
        \includegraphics[width=0.131\linewidth]{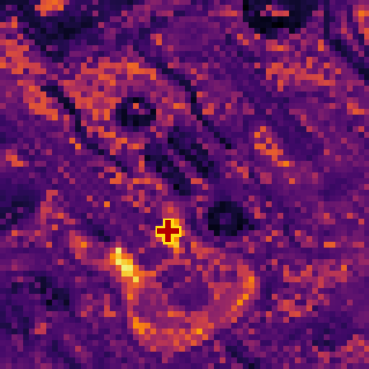} &
        \includegraphics[width=0.131\linewidth]{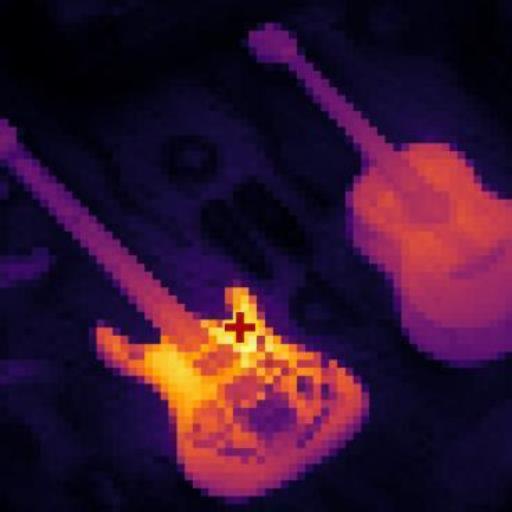} &
        \includegraphics[width=0.131\linewidth]{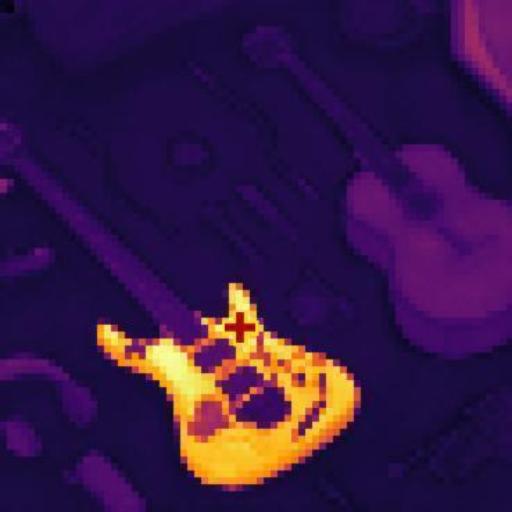} & &
        \includegraphics[width=0.131\linewidth]{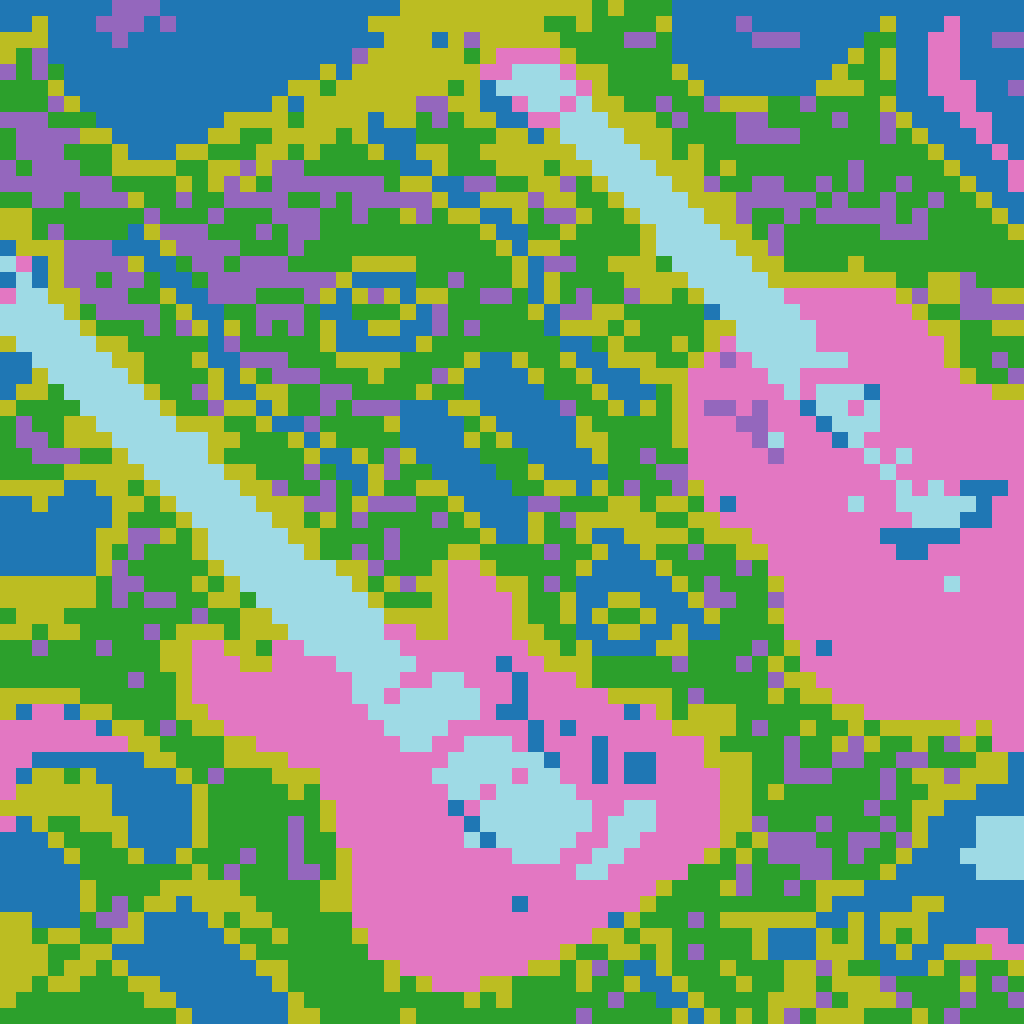} &
        \includegraphics[width=0.131\linewidth]{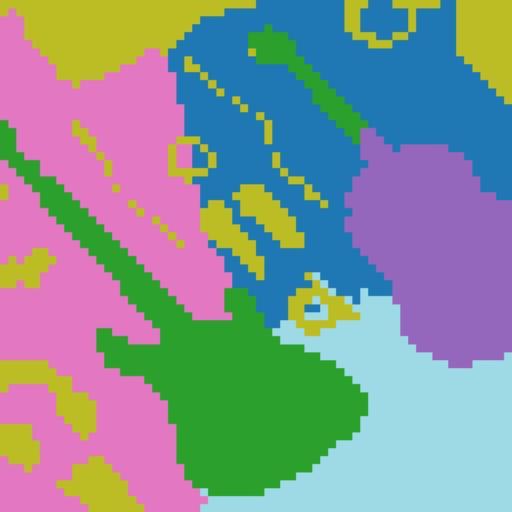} &
        \includegraphics[width=0.131\linewidth]{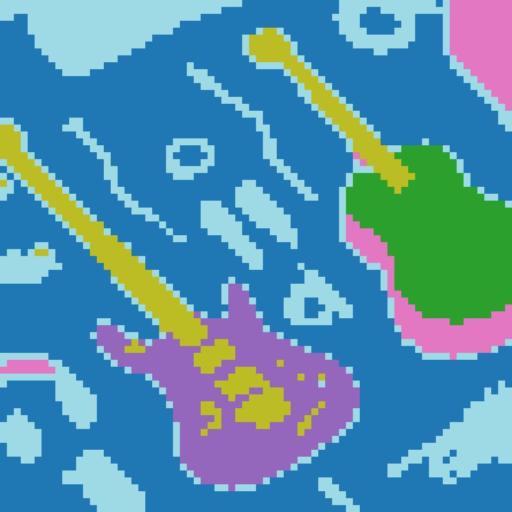} \\

        \vspace{-0.8mm}\includegraphics[width=0.131\linewidth]{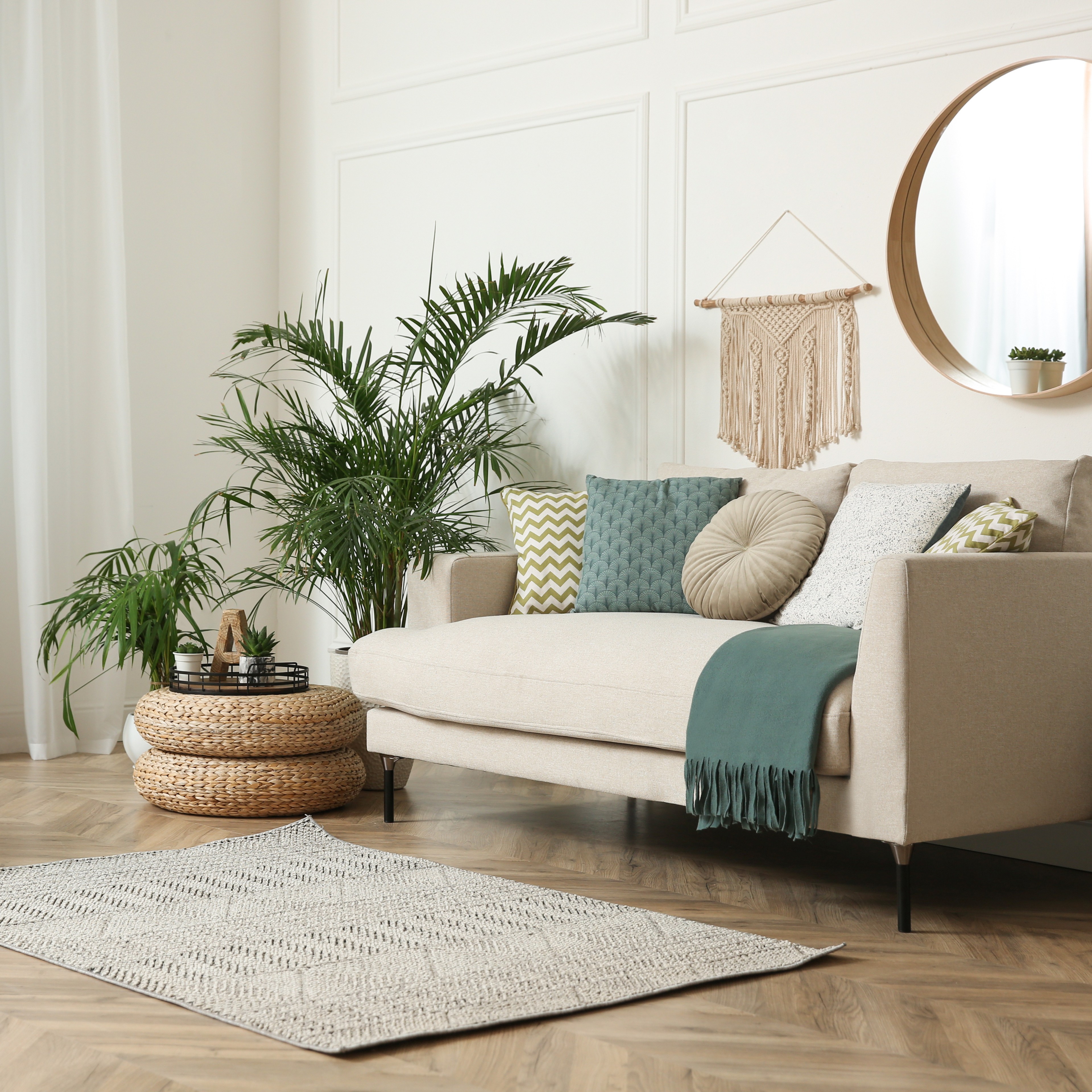} & &
        \includegraphics[width=0.131\linewidth]{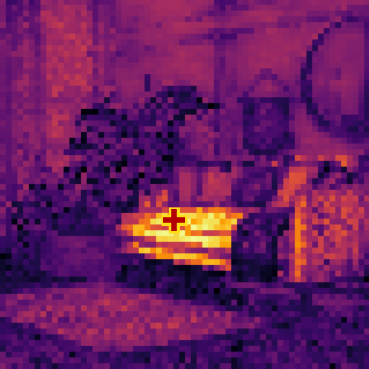} &
        \includegraphics[width=0.131\linewidth]{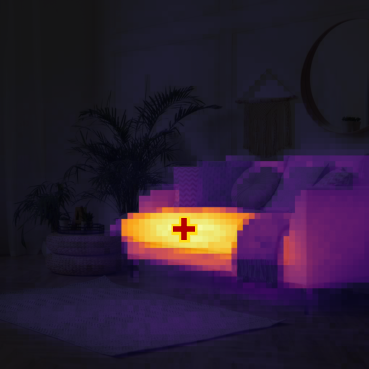} &
        \includegraphics[width=0.131\linewidth]{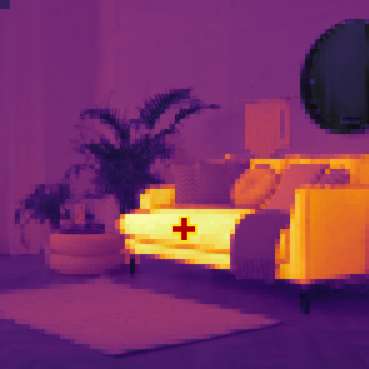} & &
        \includegraphics[width=0.131\linewidth]{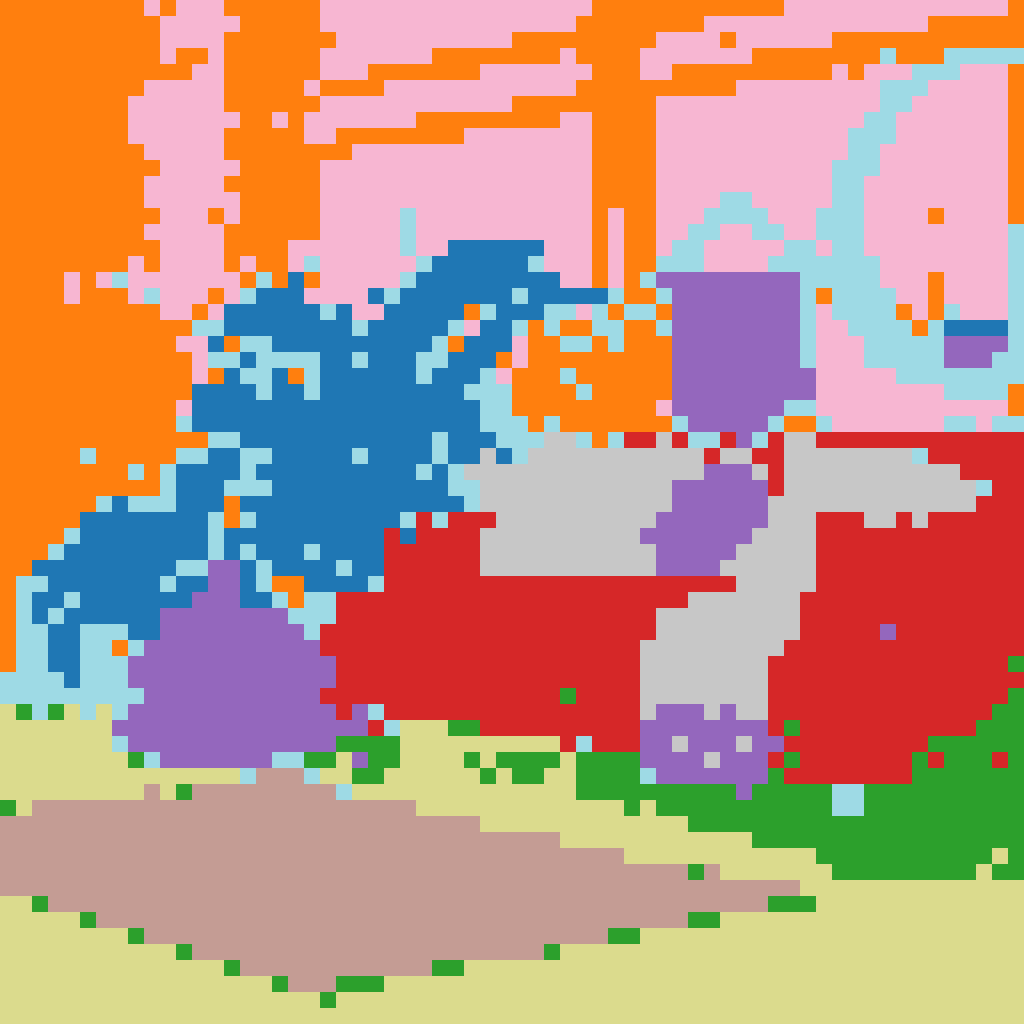} &
        \includegraphics[width=0.131\linewidth]{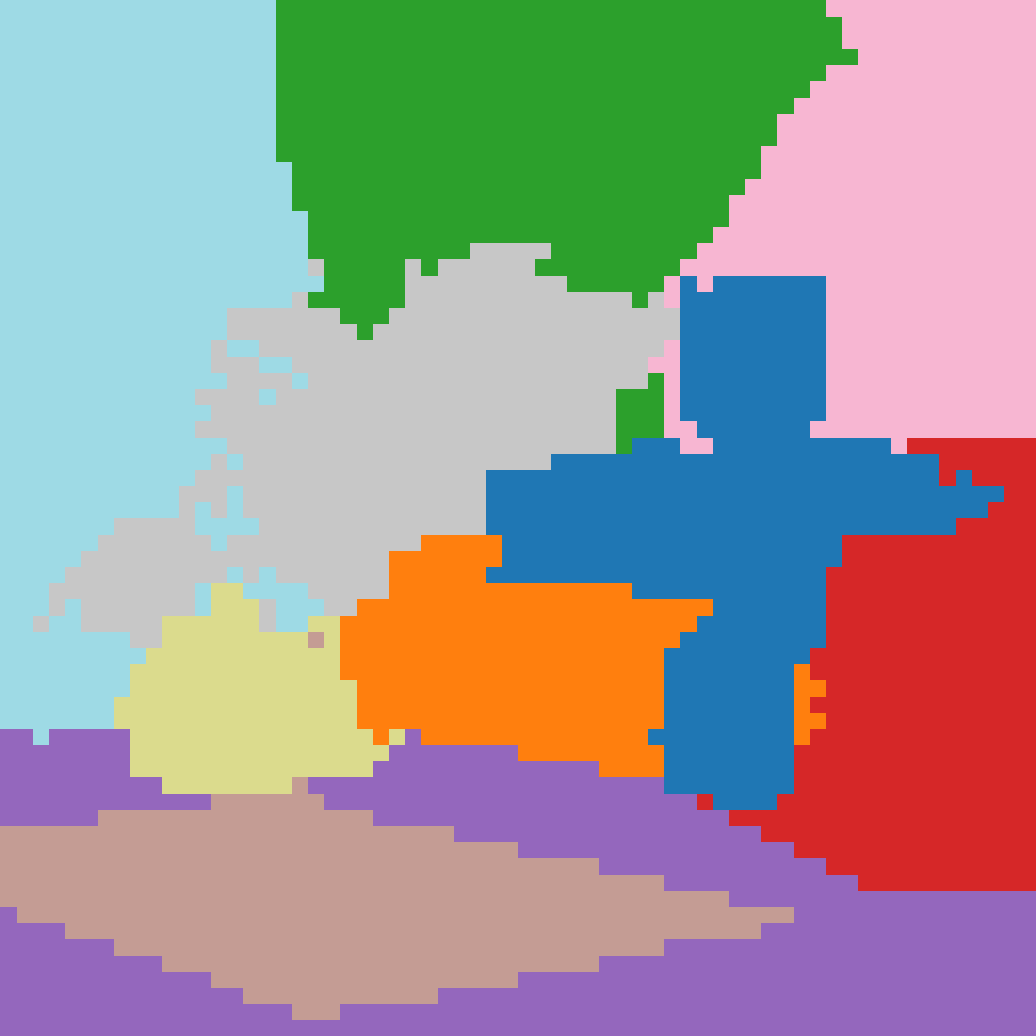} &
        \includegraphics[width=0.131\linewidth]{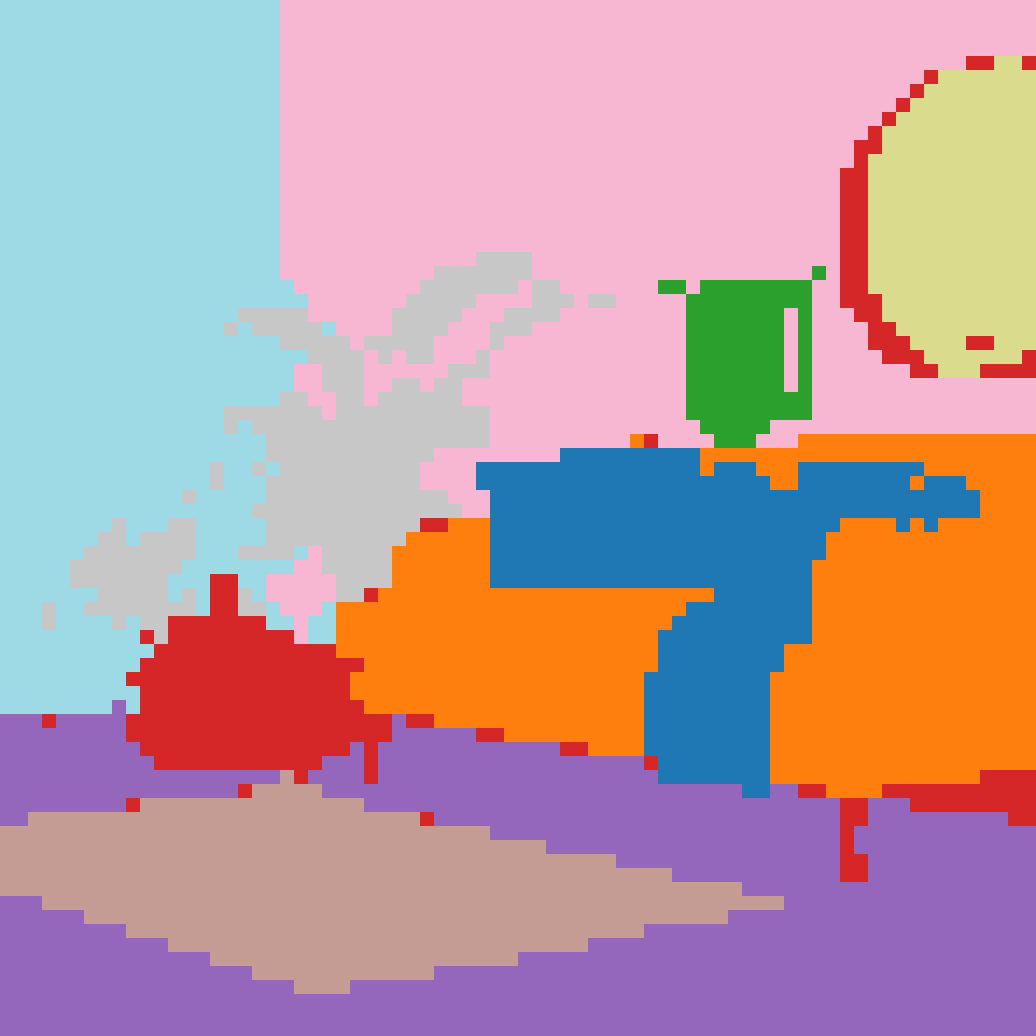} \\

    \end{tabular}

    \caption{\textbf{Patch-wise similarity and unsupervised segmentation.}
    Left group shows cosine similarity maps between the embedding of a reference patch (red cross) and all others, visualizing the spatial coherence of learned representations. The examples shown gradually evolve from a flat surface to a medium-scale scene. Right group displays K-means segmentations obtained from the patch embeddings. Compared to DINOv3 and PE Spatial, \methodName produces similarity responses and clusters that are more spatially consistent toward material properties, grouping regions by reflectance and texture rather than by semantic or geometric cues, making it suitable for appearance-related tasks.}

    \label{fig:qualitative}
\end{figure}

We evaluate the representations learned by \methodName on two downstream tasks: \textbf{material selection} and \textbf{feature separability} via k-NN. Together, these assess whether the learned features capture intrinsic appearance properties independently of geometry and lighting. Additional qualitative results, including material selection on DuMaS and patch similarity in images and videos through cross-frame feature propagation, are provided in the supplementary materials.

\myparagraph{Material selection.}
We follow the evaluation protocol and DuMaS dataset of~\cite{guerrero2025fine} to assess whether features separate materials according to appearance similarity. Given a query patch, we compute its cosine similarity to all image-patch embeddings, producing a dense similarity map. This map is thresholded at 0.5 and compared against the ground-truth material segmentation. As shown in Table~\ref{tab:selection}, \methodName consistently outperforms all generic feature baselines, including DINOv3~\cite{simeoni2025dinov3}, from which it is initialized.
Agglomerative models such as PE Spatial~\cite{bolya2025perception} and RADIOv3~\cite{heinrich2025radiov2} provide competitive dense-prediction features, but remain less aligned with physical material boundaries. Conversely, vision-language models such as CLIP~\cite{radford2021clip} and SigLIP~2~\cite{tschannen2025siglip2} perform poorly, as their embeddings tend to conflate material identity with object-level semantics. Overall, \methodName produces more coherent similarity maps that better follow material boundaries while remaining robust to changes in illumination and geometry.
We also report performance for Materialistic~\cite{sharma2023materialistic} and Guerrero et al.~\cite{guerrero2025fine} as task-specific upper bounds. Unlike these methods, which use pixel-level supervision and dedicated dense-prediction architectures, \methodName is evaluated directly from ViT patch-level backbone features, without supervised task adaptation. Thus, in downstream pipelines, \methodName is intended to replace generic backbones such as DINOv3 rather than full task-specific prediction architectures.

\myparagraph{Feature separability via k-NN evaluation.}
To assess whether the learned representations organize materials into semantically meaningful clusters, we adopt a non-parametric $k$-nearest neighbors (k-NN) evaluation. We use a controlled synthetic test set containing $972$ materials within $16$ categories. Each material is rendered under $6$ geometry templates and $4$ lighting conditions, producing $24$ distinct variants per material, resulting in a total of $23,328$ renders.

Each image is embedded with the frozen encoder. For models without a global token, such as \textit{PE Spatial}~\cite{bolya2025perception}, we use global average pooling over spatial features. For each query sample, we retrieve its $k=16$ nearest neighbors and assign a label through weighted majority voting, where weights are proportional to the cosine similarity. Classification accuracy is then computed with respect to the ground-truth material labels. As shown in Table~\ref{tab:selection}, \methodName significantly outperforms all baselines, including DINOv3. 
These results show that the physically oriented pretraining of \methodName produces tighter, more homogeneous clusters that are grounded in intrinsic material properties. Interestingly, models optimized for dense prediction or universal distillation, such as PE Spatial and RADIOv3, struggle in this global classification setting, further highlighting that \methodName captures a more robust physical material identity than semantic or multi-purpose backbones.

\begin{figure*}[t]
    \centering
    \setlength{\tabcolsep}{.5pt}
    \begin{tabular}{rrrrrrr}
        
        Input & &
        \includegraphics[width=0.167\linewidth, valign=m]{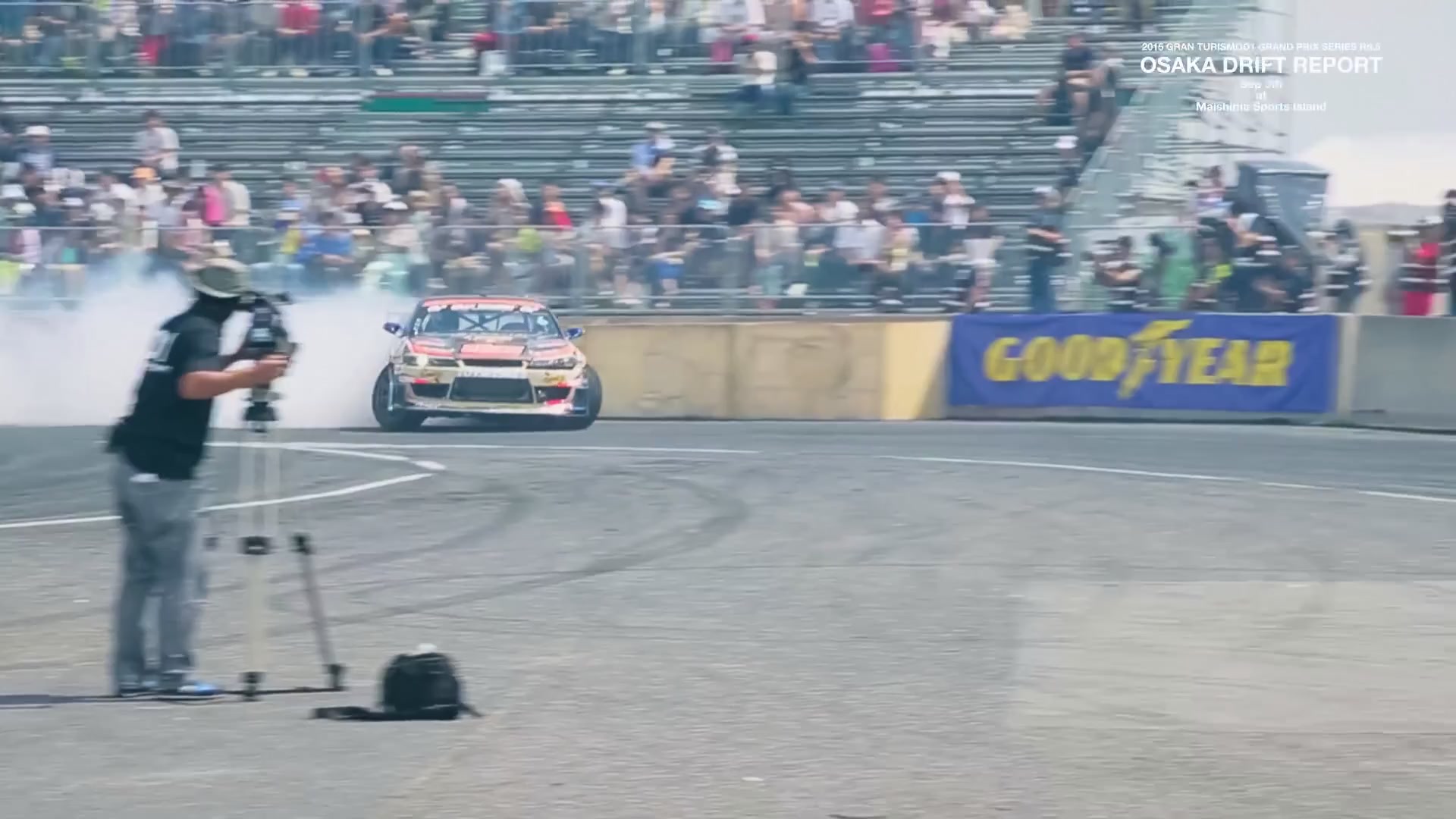} &
        \includegraphics[width=0.167\linewidth, valign=m]{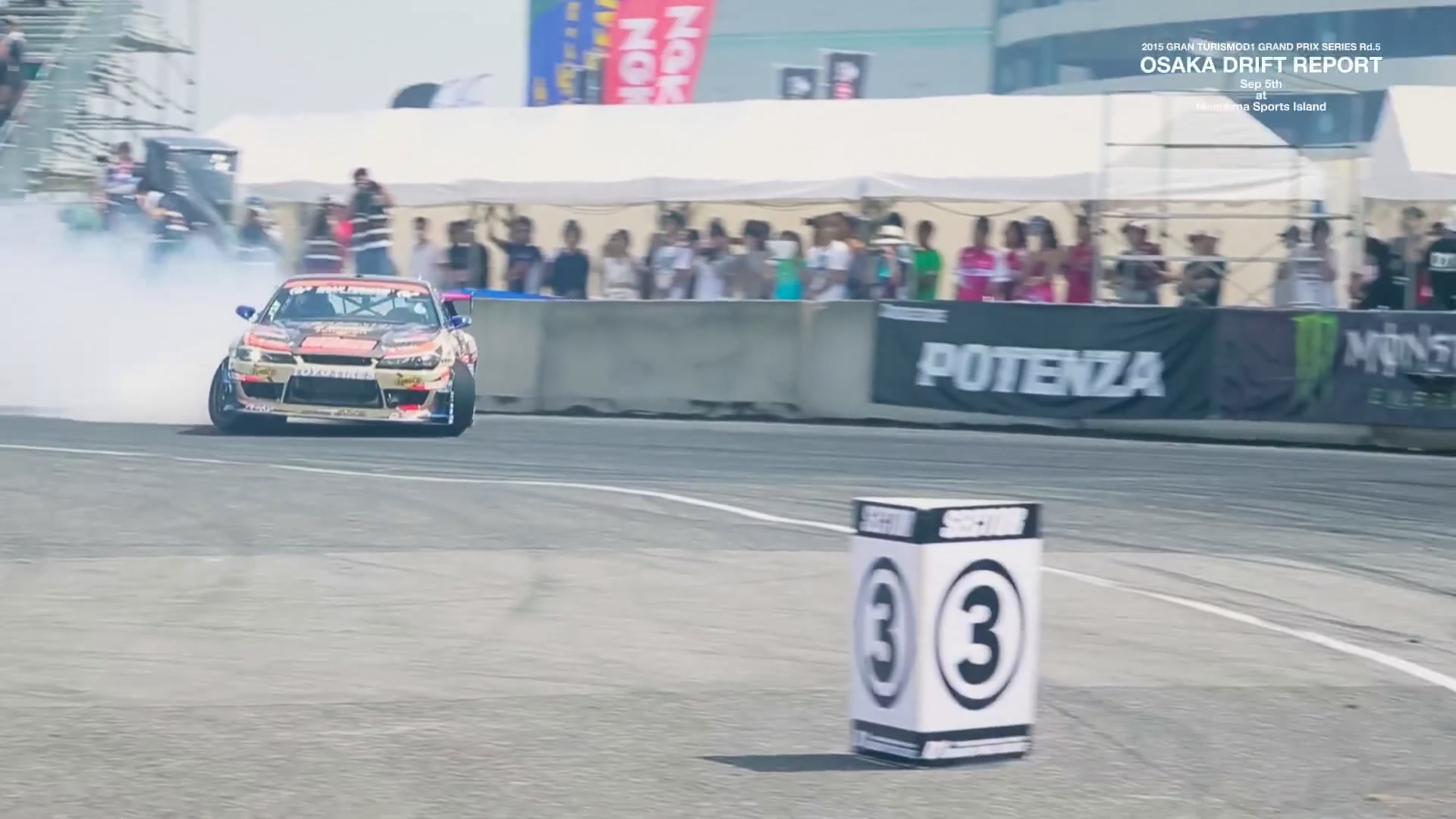} &
        \includegraphics[width=0.167\linewidth, valign=m]{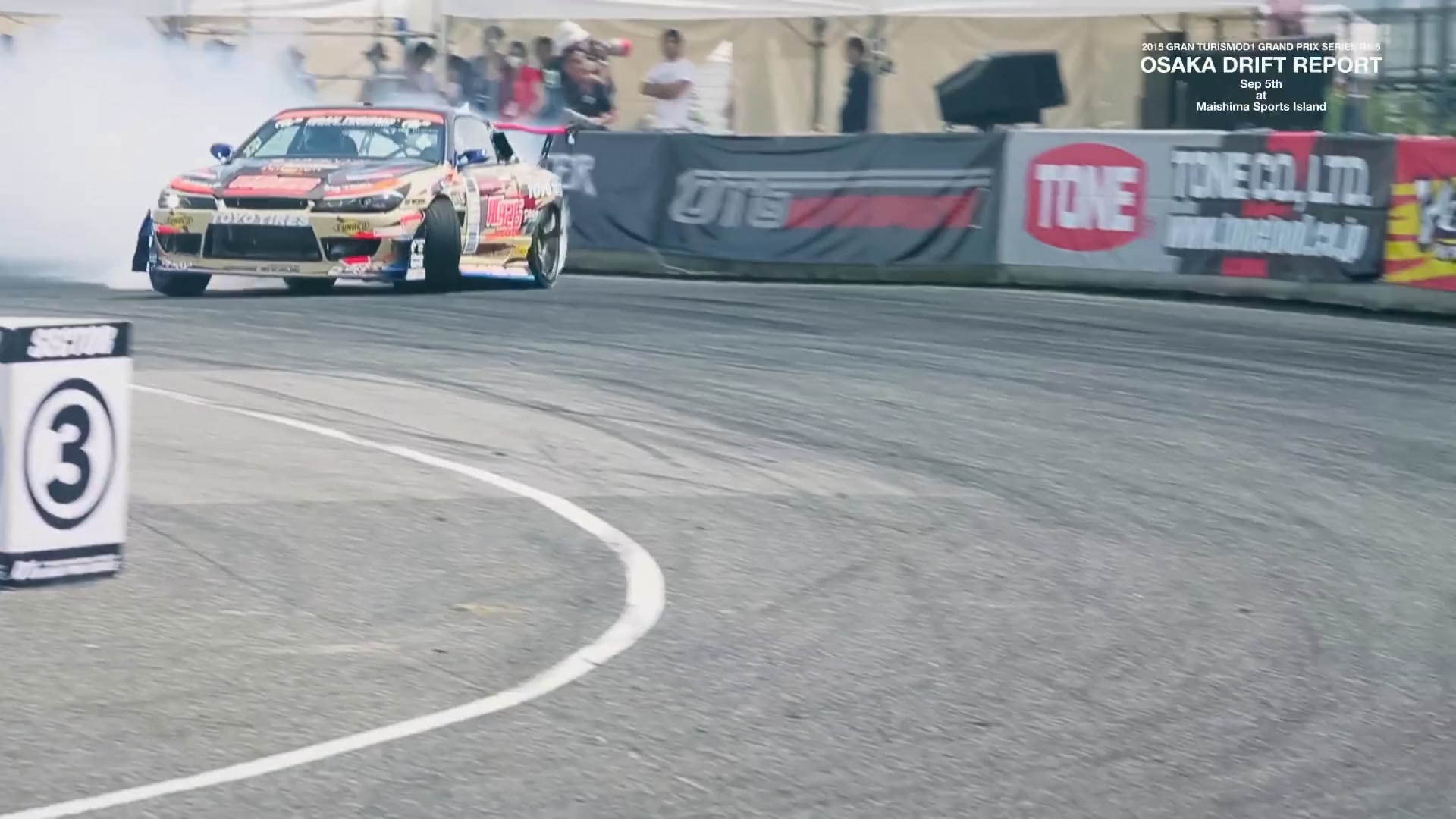} &
        \includegraphics[width=0.167\linewidth, valign=m]{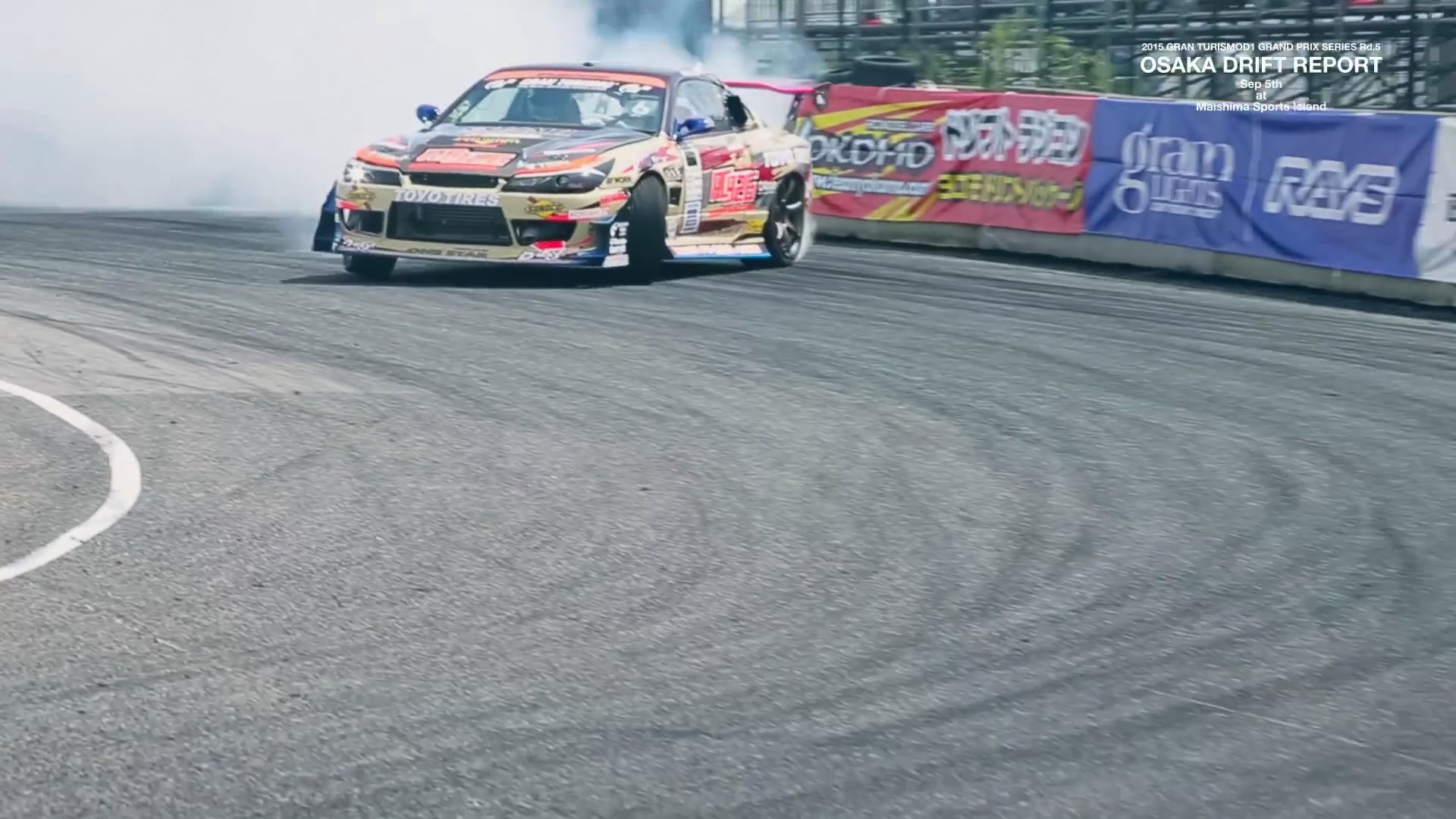} &
        \includegraphics[width=0.167\linewidth, valign=m]{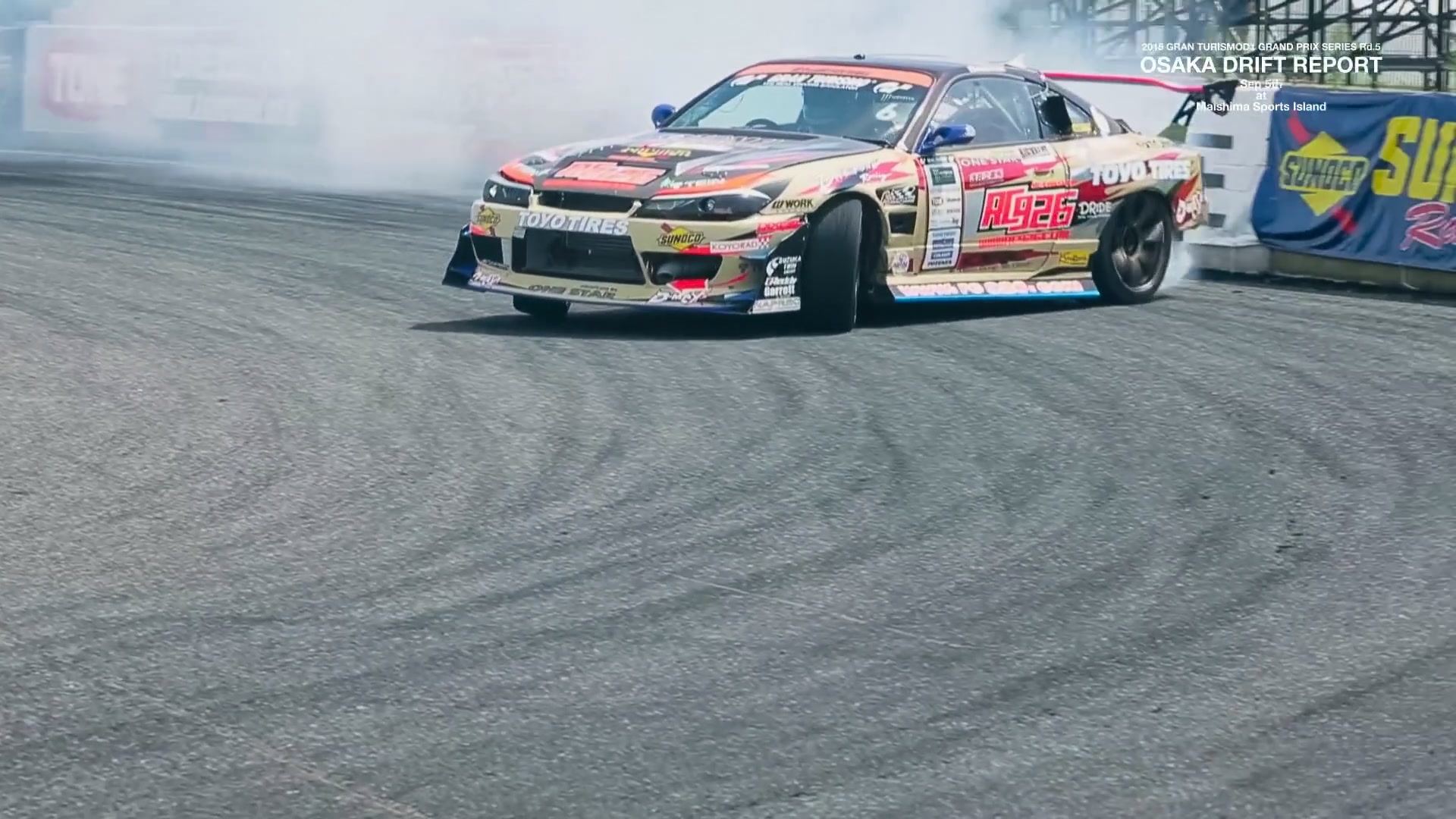} \\ [10.5pt]
        PE Spatial & \hspace{5pt} &
        \includegraphics[width=0.167\linewidth, valign=m]{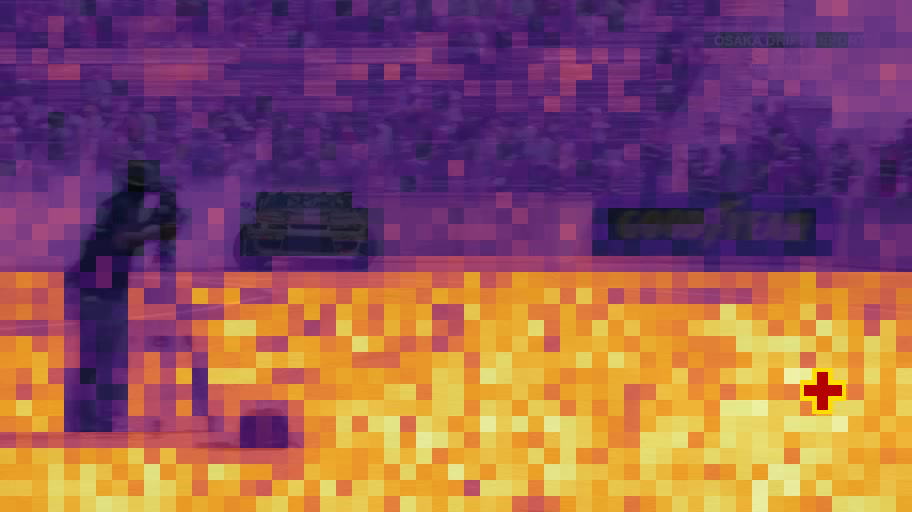} &
        \includegraphics[width=0.167\linewidth, valign=m]{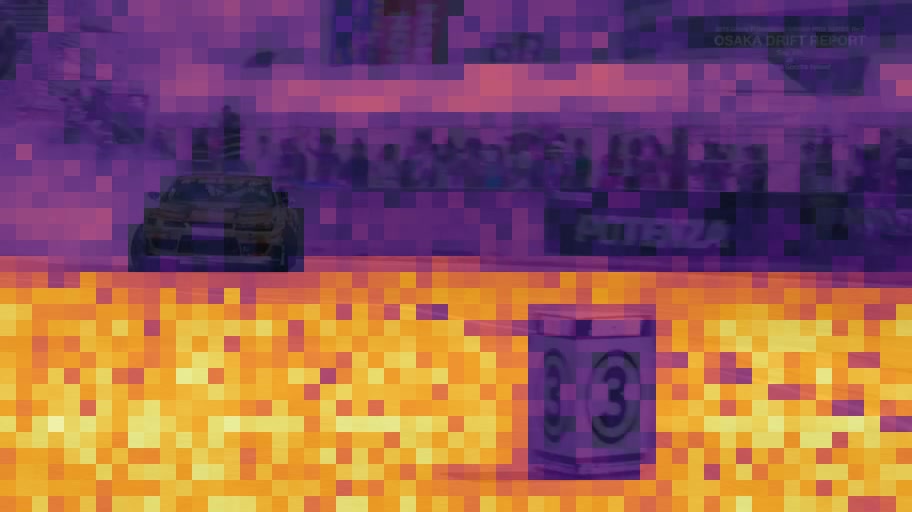} &
        \includegraphics[width=0.167\linewidth, valign=m]{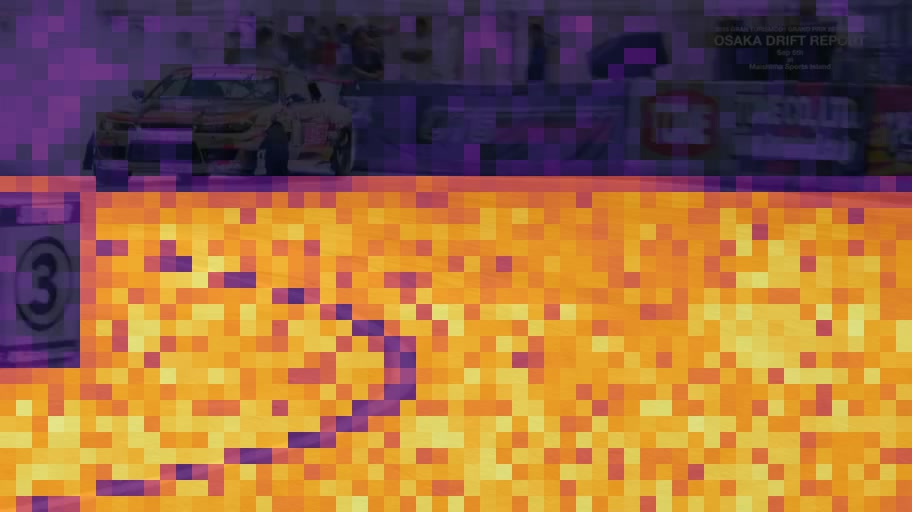} &
        \includegraphics[width=0.167\linewidth, valign=m]{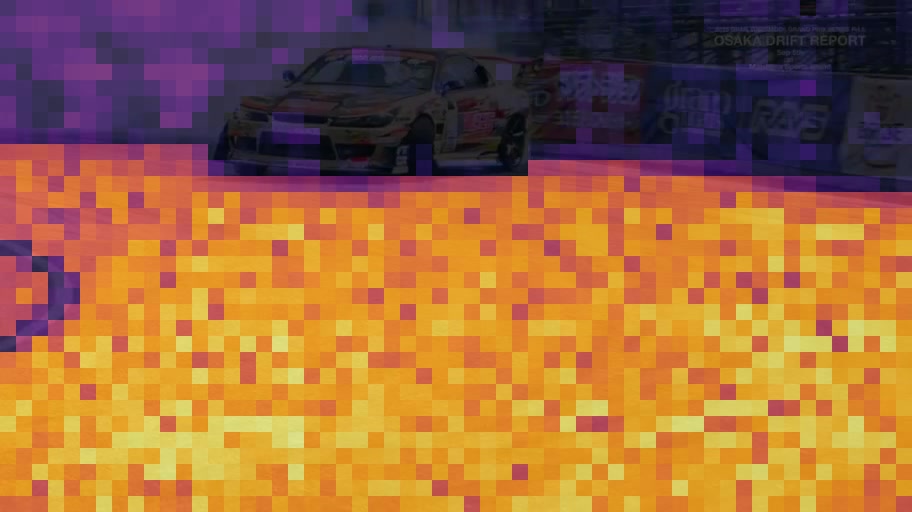} &
        \includegraphics[width=0.167\linewidth, valign=m]{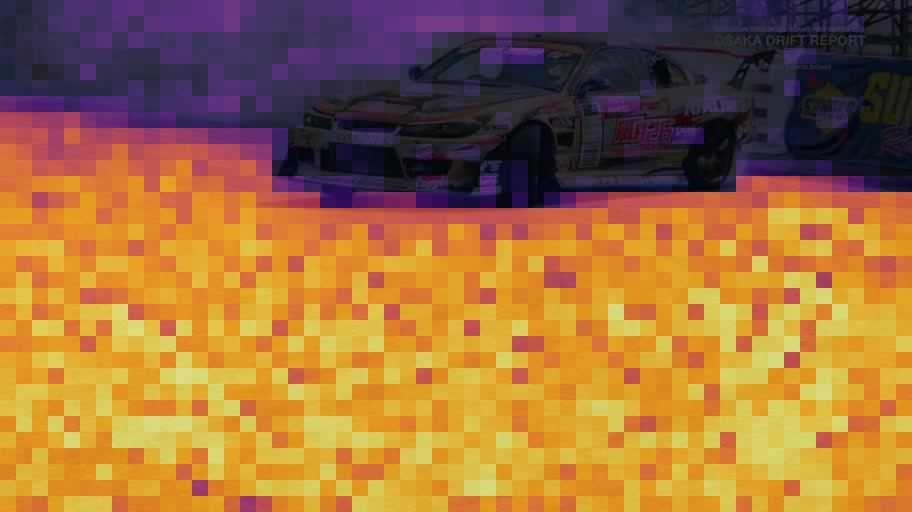} \\ [10.5pt]
        DINOv3 & \hspace{5pt} &
        \includegraphics[width=0.167\linewidth, valign=m]{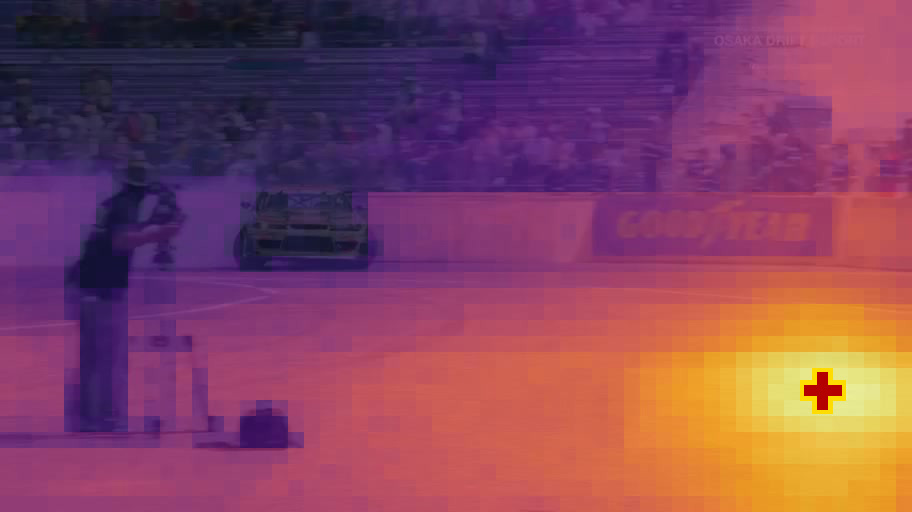} &
        \includegraphics[width=0.167\linewidth, valign=m]{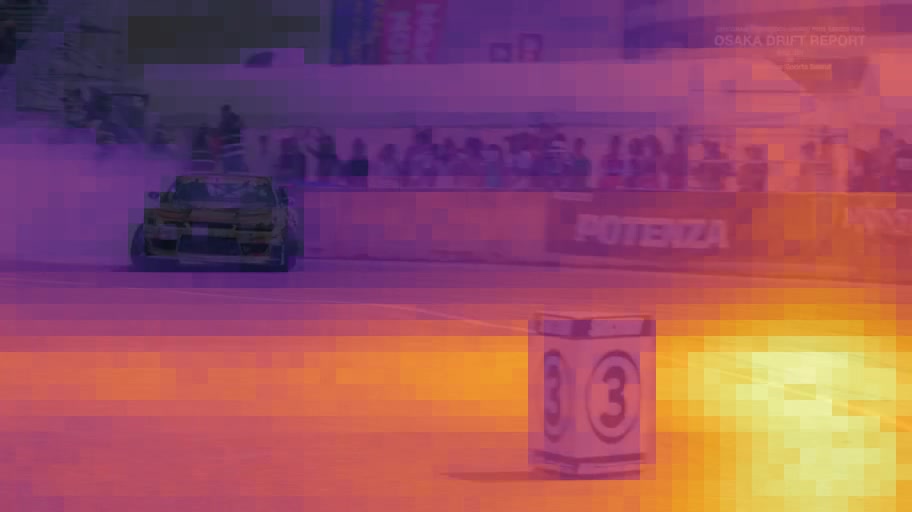} &
        \includegraphics[width=0.167\linewidth, valign=m]{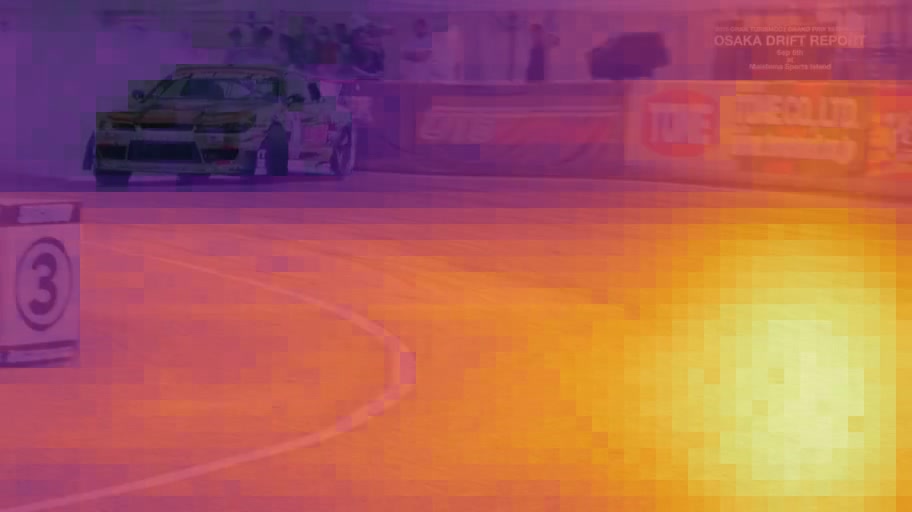} &
        \includegraphics[width=0.167\linewidth, valign=m]{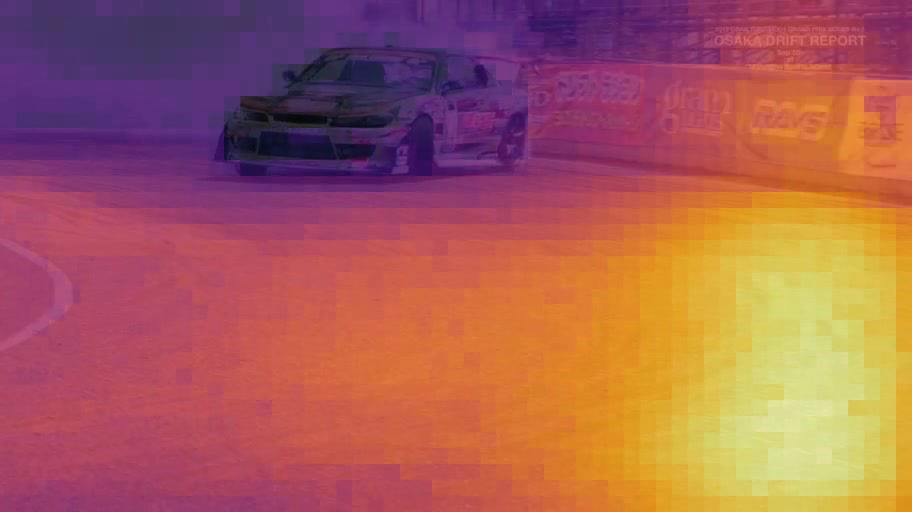} &
        \includegraphics[width=0.167\linewidth, valign=m]{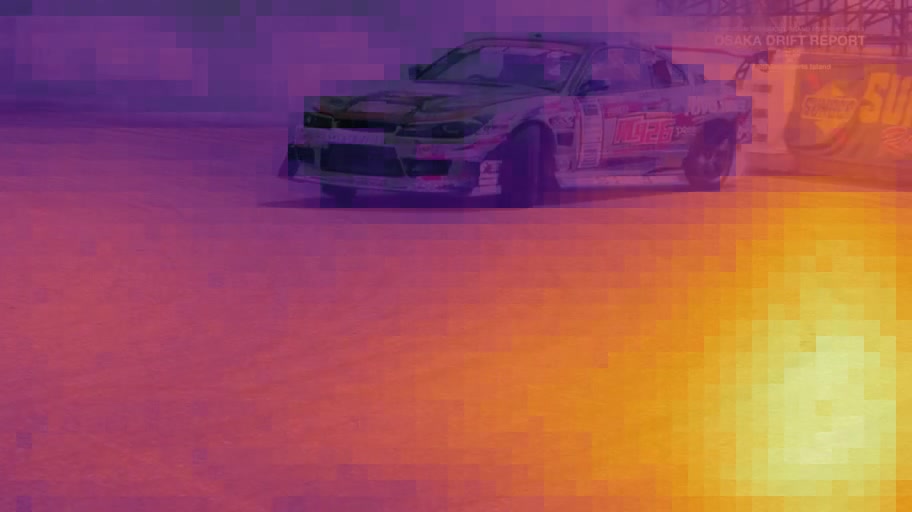} \\ [10.5pt]
        \textbf{\methodName} & \hspace{5pt} &
        \includegraphics[width=0.167\linewidth, valign=m]{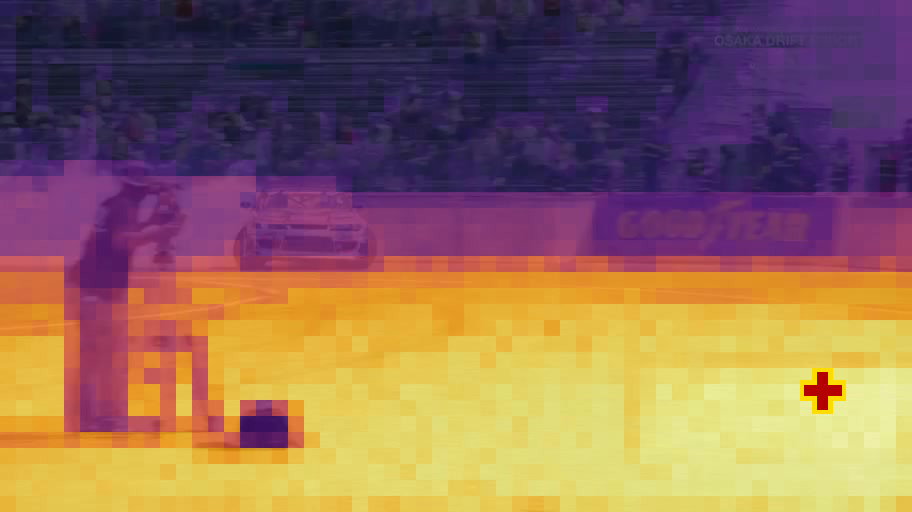} &
        \includegraphics[width=0.167\linewidth, valign=m]{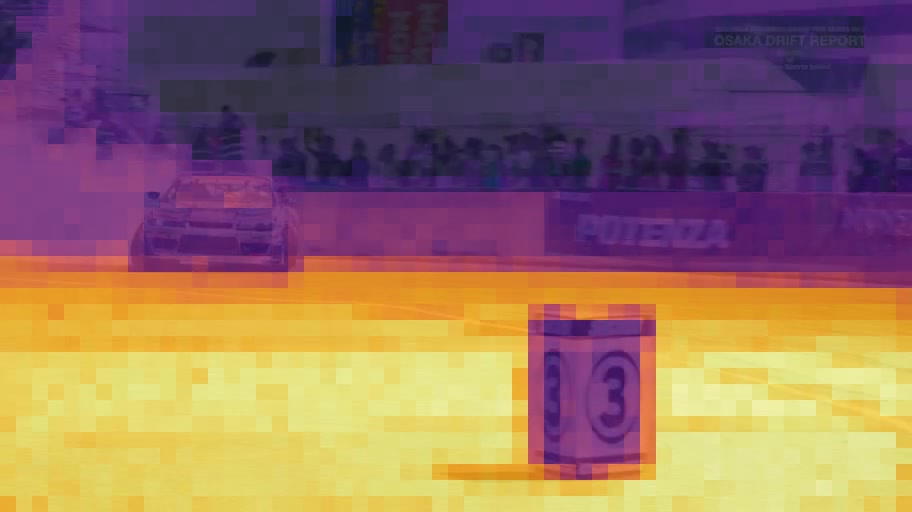} &
        \includegraphics[width=0.167\linewidth, valign=m]{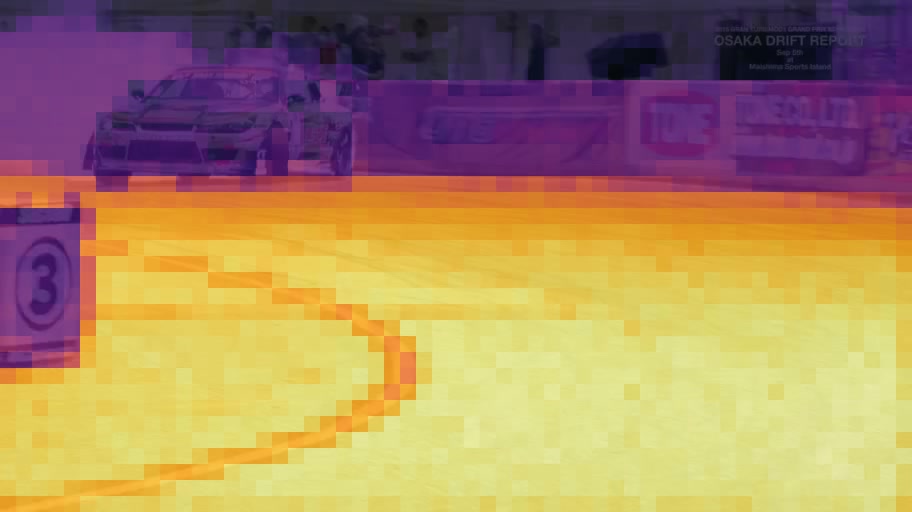} &
        \includegraphics[width=0.167\linewidth, valign=m]{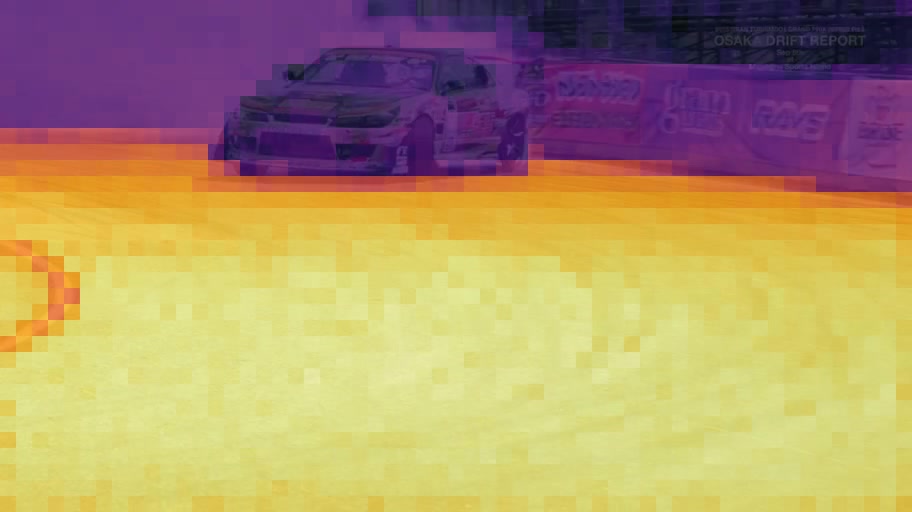} &
        \includegraphics[width=0.167\linewidth, valign=m]{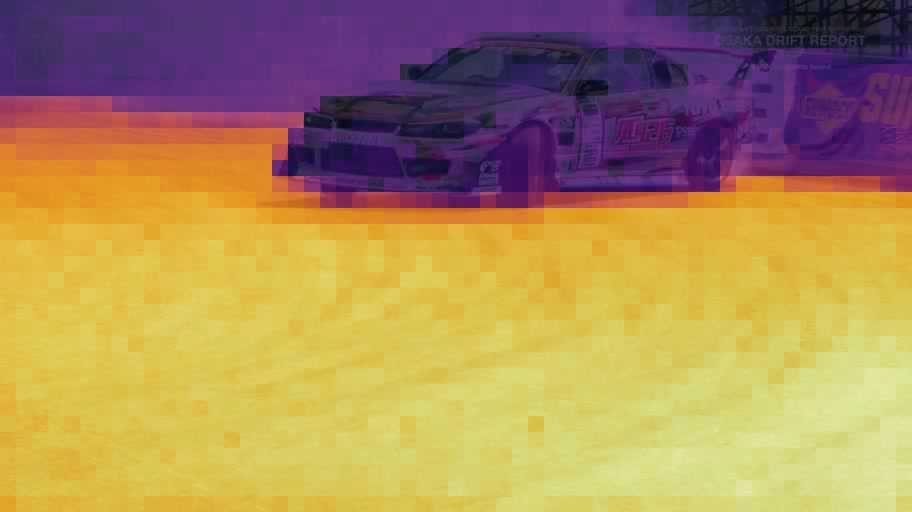} \\ [10.5pt]

        Input & &
        \includegraphics[width=0.167\linewidth, valign=m]{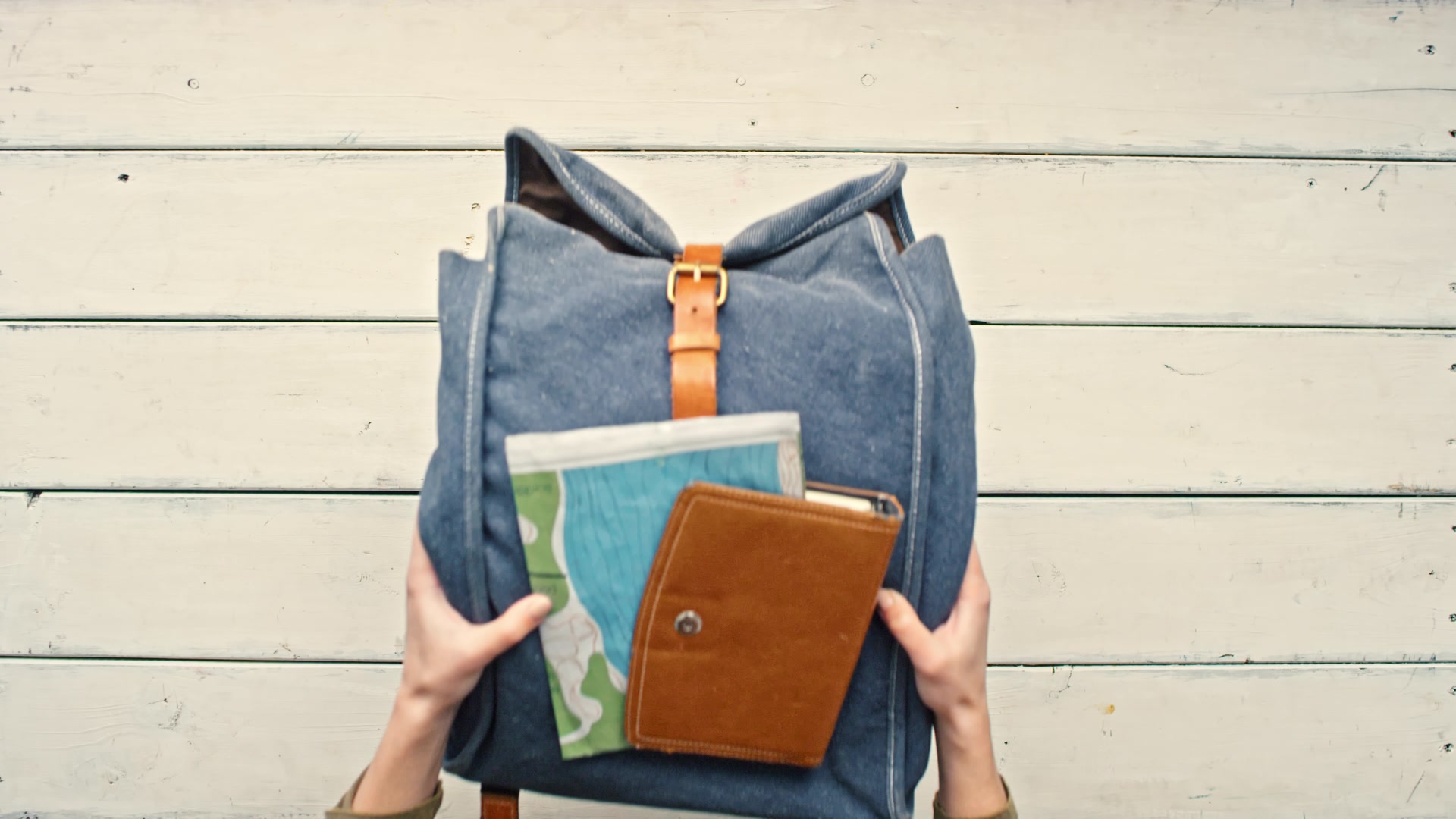} &
        \includegraphics[width=0.167\linewidth, valign=m]{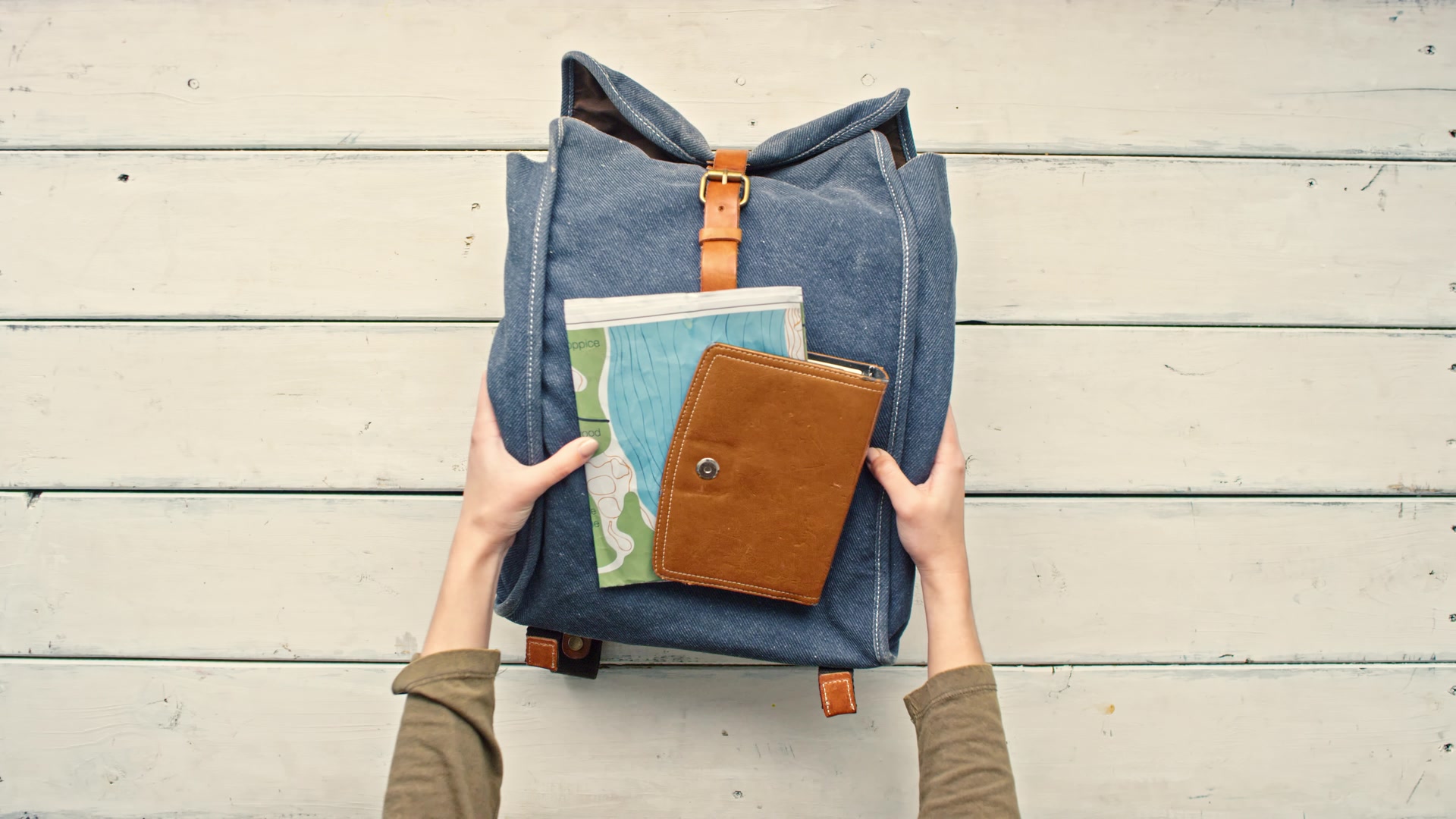} &
        \includegraphics[width=0.167\linewidth, valign=m]{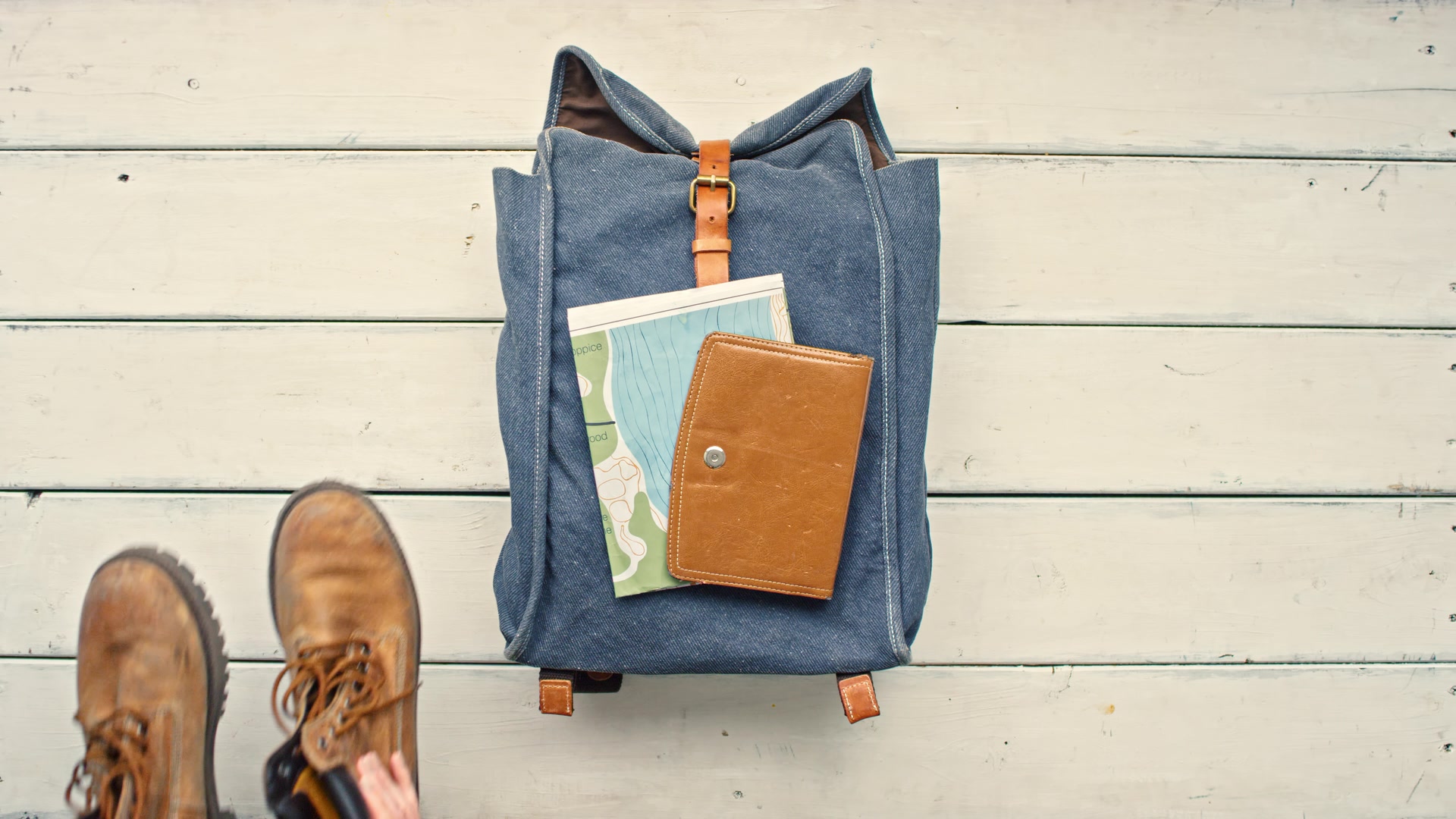} &
        \includegraphics[width=0.167\linewidth, valign=m]{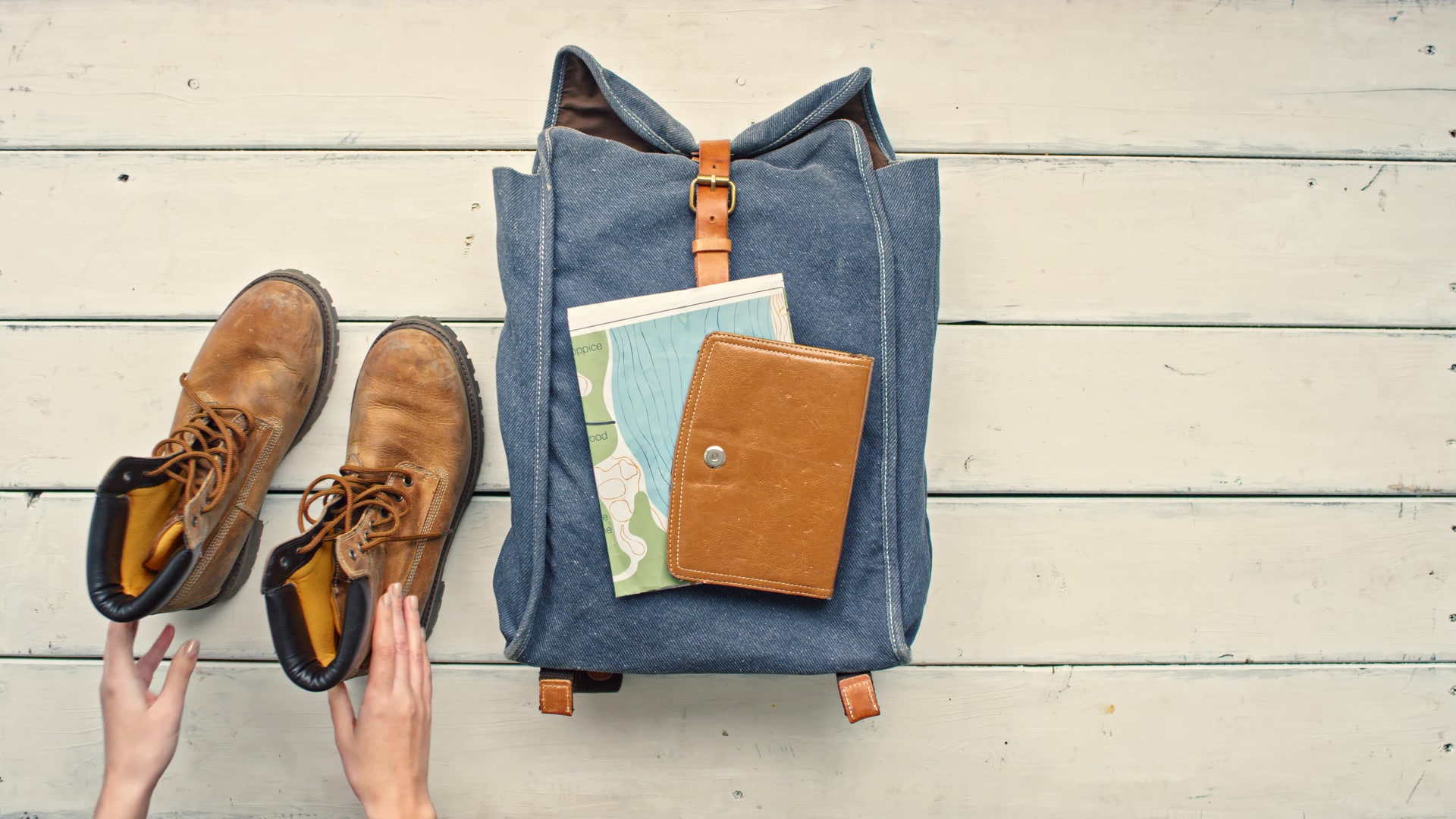} &
        \includegraphics[width=0.167\linewidth, valign=m]{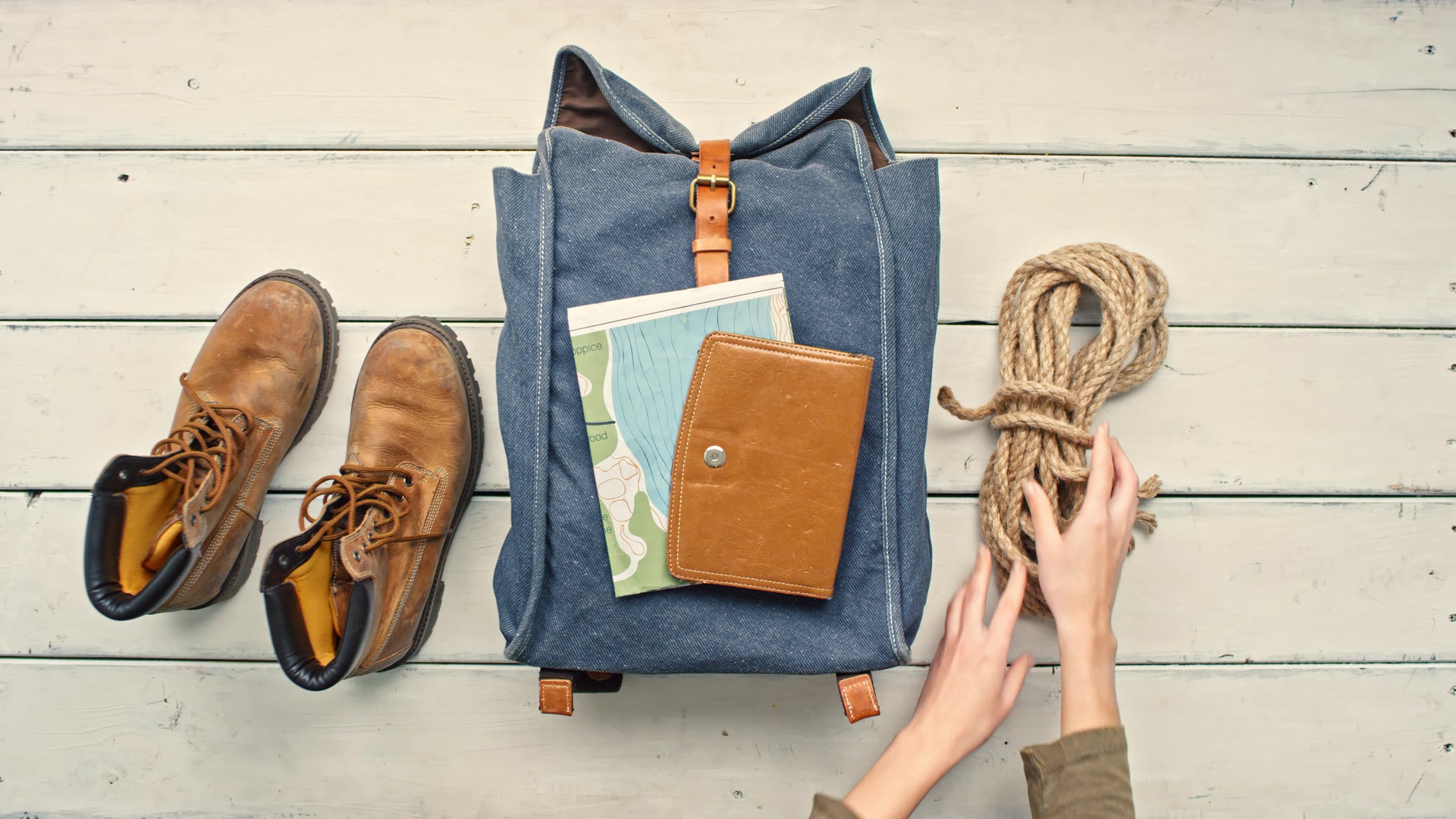} \\ [10.5pt]
        PE Spatial & \hspace{5pt} &
        \includegraphics[width=0.167\linewidth, valign=m]{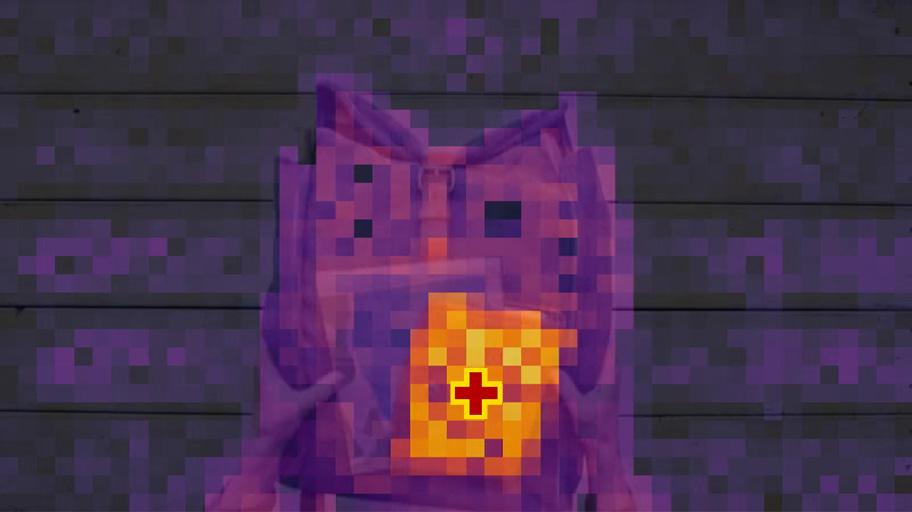} &
        \includegraphics[width=0.167\linewidth, valign=m]{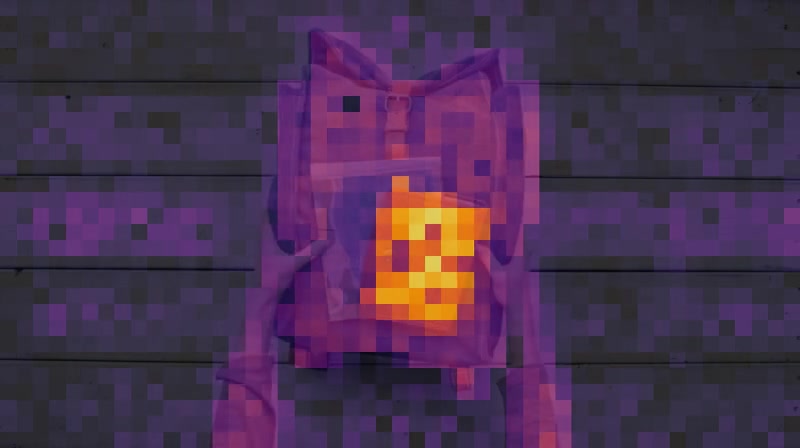} &
        \includegraphics[width=0.167\linewidth, valign=m]{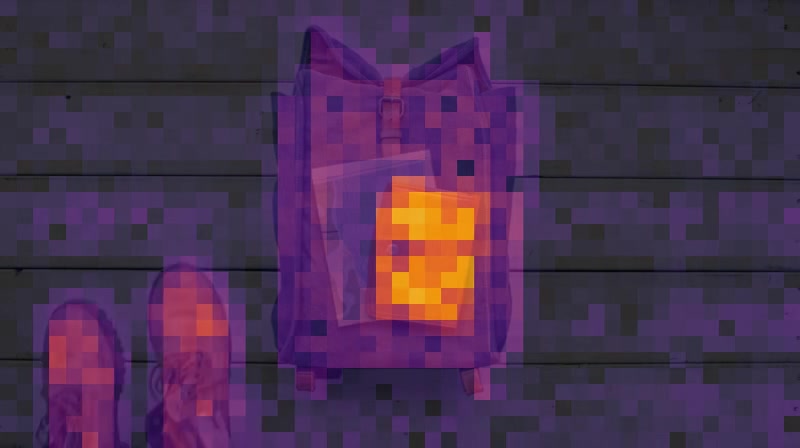} &
        \includegraphics[width=0.167\linewidth, valign=m]{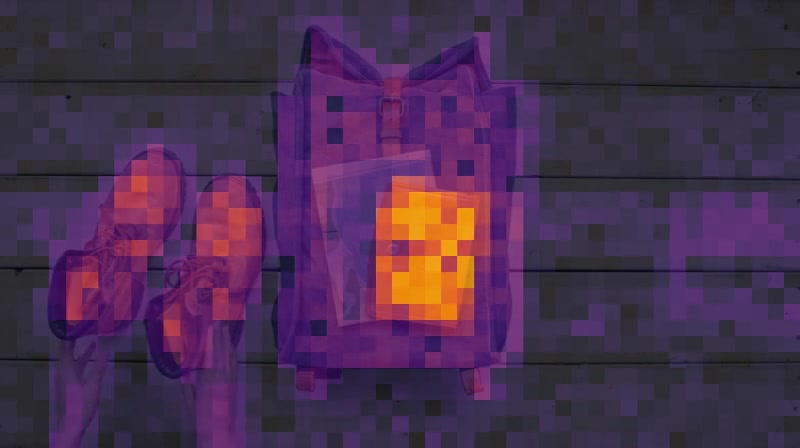} &
        \includegraphics[width=0.167\linewidth, valign=m]{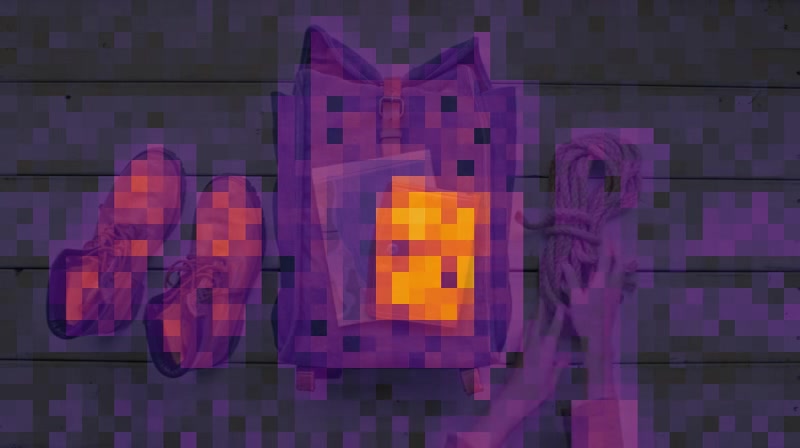} \\  [10.5pt]
        DINOv3 & \hspace{5pt} &
        \includegraphics[width=0.167\linewidth, valign=m]{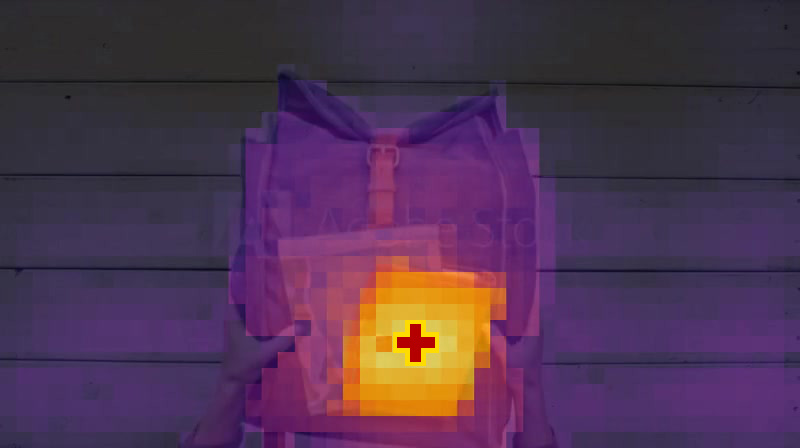} &
        \includegraphics[width=0.167\linewidth, valign=m]{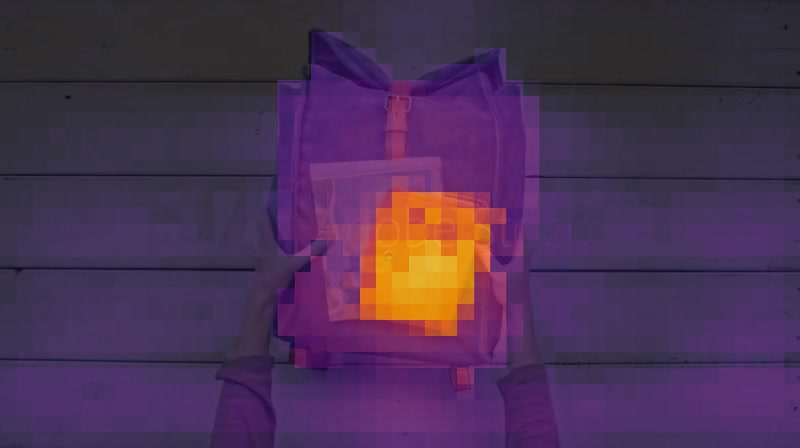} &
        \includegraphics[width=0.167\linewidth, valign=m]{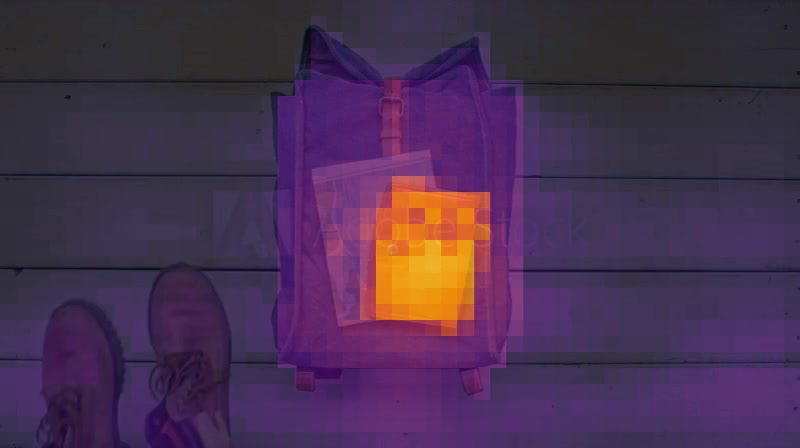} &
        \includegraphics[width=0.167\linewidth, valign=m]{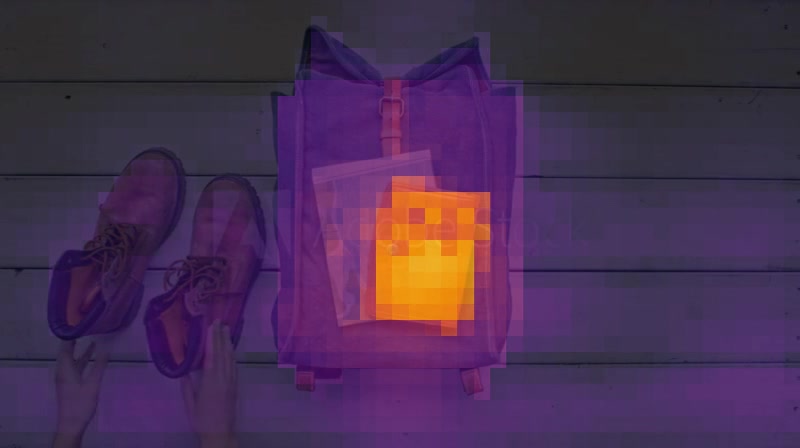} &
        \includegraphics[width=0.167\linewidth, valign=m]{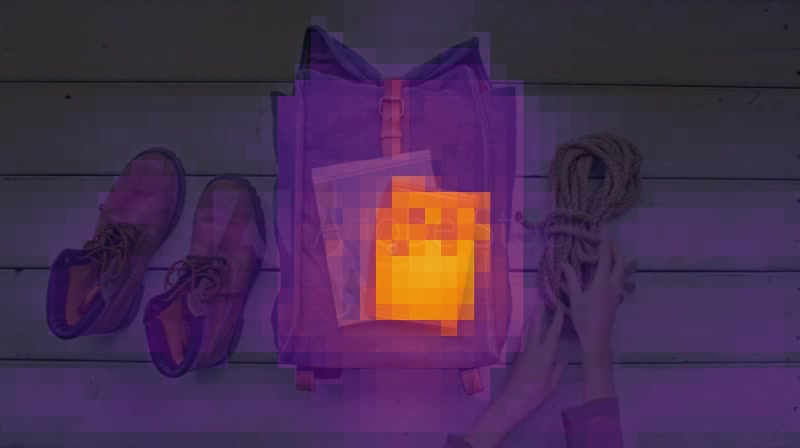} \\  [10.5pt]
        \textbf{\methodName} & \hspace{5pt} &
        \includegraphics[width=0.167\linewidth, valign=m]{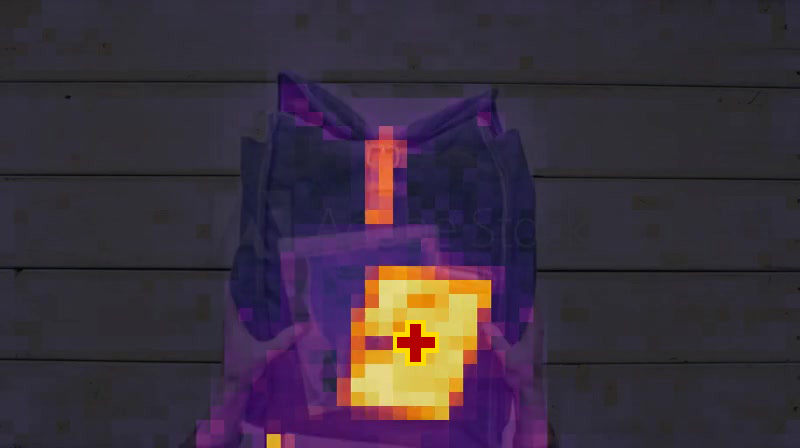} &
        \includegraphics[width=0.167\linewidth, valign=m]{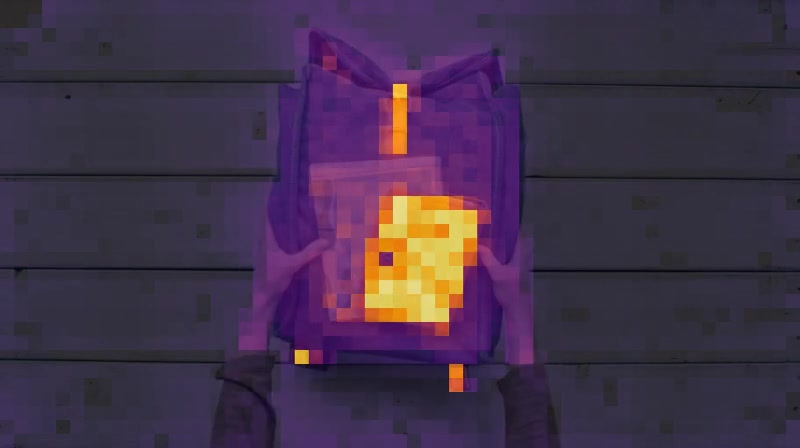} &
        \includegraphics[width=0.167\linewidth, valign=m]{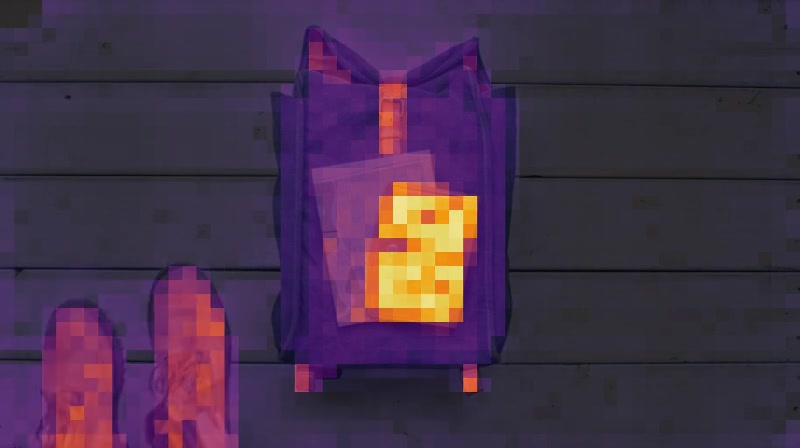} &
        \includegraphics[width=0.167\linewidth, valign=m]{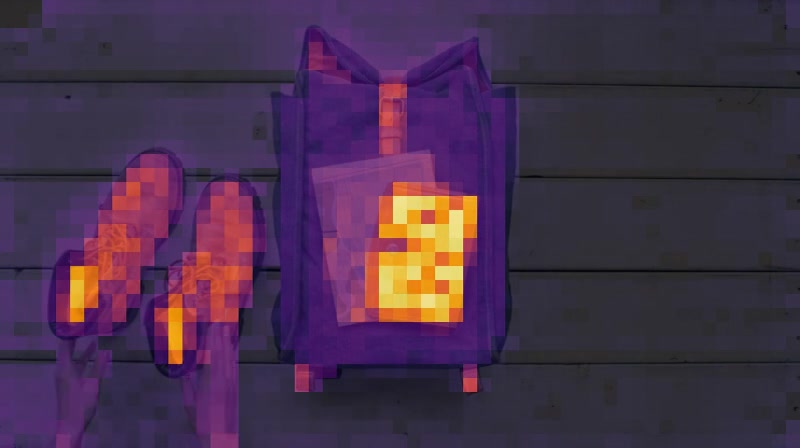} &
        \includegraphics[width=0.167\linewidth, valign=m]{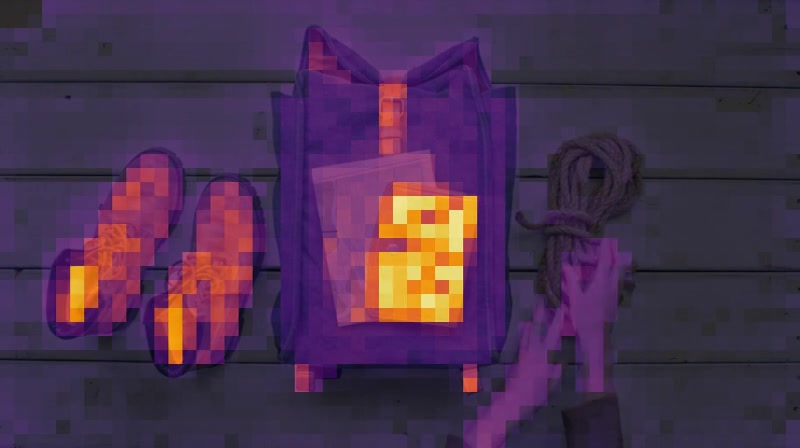} \\
    \end{tabular}

    \caption{\textbf{Feature similarity in videos}. The selection is initialized in the first frame and propagated throughout the video. We compute the cosine similarity maps between the embedding of a reference patch (red cross) and all patches of all frames in the video. Over time, \methodName maintains better coherence and alignment with material boundaries compared to DINOv3 and PE Spatial. In the first video, \methodName more uniformly selects the asphalt compare to the other methods, while in the second it identifies all the leather parts regardless of the object semantics.}
    \label{fig:video}
\end{figure*}

\myparagraph{Robustness.}
To assess invariance to extrinsic factors, we measure robustness across variations in illumination and geometry. For each material class we compute predictions using the same k-NN classifier described above. We then evaluate the average pairwise Hamming distance between predictions obtained under different lighting conditions (fixed geometry) and under different geometry templates (fixed lighting). Lower values indicate higher invariance. Table~\ref{tab:robustness} shows that \methodName outperforms the other methods, exhibiting less variability under both types of perturbation and confirming that the representation remains stable across extrinsic appearance changes.

\subsection{Semantic Performance}
\label{sec:semantic_performance}

We quantify the tradeoff between material alignment and general-purpose utility in \cref{tab:semantic_material_tradeoff} by evaluating \methodName on the original semantic tasks via linear probing: ImageNet classification and ADE20K/Cityscapes segmentation. Compared to DINOv3, \methodName incurs a moderate drop of 3.4 accuracy points for classification and 5.3/5.4 mIoU points for segmentation, respectively. This indicates that the material-centric adaptation \textbf{preserves much of DINOv3’s semantic} utility while reorienting the representation toward material identity. We therefore view this as a \textbf{specialization tradeoff}: DINOv3 remains stronger for purely semantic tasks, while \methodName provides more material-grounded features for material-consistency and appearance-based tasks.

\begin{table}[ht]
\centering
\caption{\textbf{Semantic--material tradeoff vs. DINOv3.} We evaluate the semantic performance retained by \methodName after material-centric adaptation using linear probing on ImageNet classification (acc, $\uparrow$) and ADE20K/Cityscapes semantic segmentation (mIoU, $\uparrow$). Although \methodName incurs a moderate drop compared to the original DINOv3 backbone, it preserves most of its semantic utility while reorienting the representation toward material-aware features.}

\label{tab:semantic_material_tradeoff}
\begin{tabular}{l@{\hspace{4pt}}ccc}
\toprule
& \textbf{Classification} & \multicolumn{2}{c}{\textbf{Semantic Segmentation}} \\
\cmidrule(lr){2-2}\cmidrule(lr){3-4}
& ImageNet & ADE20K & Cityscapes \\
\midrule
& Acc. $\uparrow$ & \multicolumn{2}{c}{mIoU $\uparrow$} \\
\midrule
DINOv3 & 83.5 & 51.5 & 71.5 \\
\methodName{} (\textbf{ours}) & 80.1 & 46.2 & 66.1 \\
\bottomrule
\end{tabular}
\end{table}

\subsection{Qualitative Results}
\label{sec:qual_results}

We qualitatively compare \methodName against DINOv3~\cite{simeoni2025dinov3} and PE Spatial~\cite{bolya2025perception} using patch-wise similarity and unsupervised segmentation (Fig.~\ref{fig:qualitative}). Each heatmap shows cosine similarity between a reference patch, marked with a cross, and all other image patches. DINOv3 often highlights semantically or spatially related regions, typically parts of the same object, even when materials differ, while PE Spatial produces noisier responses. In contrast, \methodName yields coherent similarity maps aligned with material boundaries, grouping patches by reflectance and texture rather than object identity. Fig.~\ref{fig:video} shows cross-frame selection in videos, where the selection is initialized in the first frame and propagated over time. Compared to DINOv3 and PE Spatial, \methodName better preserves coherence and alignment with material boundaries across frames.

For unsupervised segmentation, we apply K-means clustering to patch embeddings to visualize how features organize the image space. The number of clusters is selected automatically by evaluating $K \in \{2,\ldots,12\}$ and choosing the value that maximizes the silhouette score~\cite{rousseeuw1987silhouettes}. Segmentations from DINOv3 and PE Spatial often follow semantic cues, grouping object parts while mixing distinct materials. \methodName instead produces spatially consistent and physically meaningful clusters that better separate regions according to material properties.

These qualitative results show that \methodName learns representations driven by reflectance and texture rather than semantics, capturing physically grounded appearance features robust to geometry and lighting.
Additional results showing invariance to lighting are provided in the supplementary materials.

\newcolumntype{L}{>{\raggedright\arraybackslash}X}
\newcolumntype{R}{>{\raggedleft\arraybackslash}X}

\begin{table}[t]
\centering
\begin{minipage}[t]{0.46\linewidth}
    \centering
    \setlength{\tabcolsep}{1pt}
    \captionsetup{type=table}
    \caption{\textbf{Robustness to illumination and geometry variations},
    measured as the avg. pairwise Hamming distance between k-NN predictions across renderings of the same material under different lighting or geometry.
    Lower values indicate higher invariance.}
    \label{tab:robustness}
    \begin{tabularx}{\linewidth}{l@{\hspace{0pt}}R@{\hspace{20pt}}R}
        \toprule
        & \textbf{Illumination} & \textbf{Geometry} \\
        & \scriptsize{hamming} $\downarrow$ & \scriptsize{hamming} $\downarrow$ \\
        \midrule
        CLIP \cite{radford2021clip} & 0.403 & 0.534 \\
        DINOv2 \cite{oquab2023dinov2} & 0.284 & 0.435 \\
        DINOv3 \cite{simeoni2025dinov3} & 0.240 & 0.365 \\
        \midrule
        \textbf{\methodName (ours)} & \textbf{0.221} & \textbf{0.305} \\
        \bottomrule
    \end{tabularx}
\end{minipage}
\hfill
\begin{minipage}[t]{0.51\linewidth}
    \centering
    \captionsetup{type=table}
    \caption{\textbf{Ablation study.}
    Effect of each training component of \methodName{} starting from the DINOv3 backbone.
    Multi-render supervision and the contrastive objective progressively improve both material selection and k-NN classification performance, leading to stronger material grouping.}
    \label{tab:ablation}
    \begin{tabularx}{\linewidth}{l@{\hspace{-1pt}}RRR|R}
        \toprule
        & \multicolumn{3}{c}{\textbf{Material selection}} & \textbf{k-NN} \\
        & $\ell_1 \downarrow$ & IoU $\uparrow$ & $F_1 \uparrow$ & Acc. $\uparrow$ \\
        \midrule
        DINOv3 & 27.6 & 59.7 & 72.2 & 66.9 \\
        \midrule
        + single render & 26.5 & 69.4 & 78.1 & 34.5 \\
        + multi render  & 26.1 & 69.9 & 79.2 & 51.3 \\
        + contrastive   & \textbf{25.4} & \textbf{72.0} & \textbf{80.8} & \textbf{71.8} \\
        \bottomrule
    \end{tabularx}
\end{minipage}
\end{table}

\subsection{Ablation Study}
\label{sec:ablation}

We conduct an ablation study to evaluate the effect of each novel training component.
Table~\ref{tab:ablation} shows that fine-tuning with single-render material supervision already improves the material-selection metrics ($\ell_1$, IoU, $F_1$), but it also substantially reduces k-NN accuracy, showing that the model becomes more physically grounded while losing global discriminability. 
Adding the \textit{multi-render} scheme improves both material-selection and classification performance, suggesting that exposure to varied viewpoints and illumination stabilizes the representation. However, only the contrastive term fully resolves the loss of separability introduced by material supervision, enforcing compact intra-material clusters and clearer inter-material boundaries. This combination yields the strongest performance across all metrics and consistently surpasses DINOv3, leading to a material representation that is both physically grounded and discriminative.

\section{Limitations}
\label{sec:limitation}

While \methodName successfully learns invariance to extrinsic factors, it does not explicitly disentangle the latent space into interpretable physical factors. A more expressive formulation separating roughness, metalness, or albedo would allow users to manipulate features along specific physical dimensions. Moreover, our pretraining relies entirely on synthetic data; closing the domain gap by integrating unlabelled real-world data through domain adaptation or self-training remains an open challenge. As a result of its material-centric adaptation, \methodName also incurs a moderate loss on general-purpose semantic tasks compared to the original backbone, reflecting a specialization tradeoff between semantic utility and material alignment. Finally, our current physical model is primarily optimized for surface reflectance and shading. Modeling more complex light-transport phenomena, such as translucency and subsurface scattering, could further improve its physical fidelity.

\section{Conclusion}
\label{sec:conclusion}

We proposed \methodName, a weakly supervised visual backbone that learns physically grounded representations of appearance directly from unlabelled data. By replacing photometric augmentations with physically meaningful variations \textit{i.e.}, different renderings of the same material under diverse geometries and lighting conditions, our approach bridges the gap between semantic invariance and physical material understanding. Through a combination of DINO-based teacher–student alignment and contrastive regularization, \methodName captures material identity while remaining invariant to extrinsic factors such as shape and illumination.
We demonstrated that the learned features transfer effectively to downstream material tasks, including fine-grained material selection, unsupervised segmentation and clustering under changing illumination and geometry. These results confirm that material-aware pretraining can yield representations that are both robust and interpretable, without relying on semantic labels.

\bibliographystyle{splncs04}
\bibliography{bibliography}
\end{document}